\documentclass[journal, twoside]{IEEEtran}

\IEEEoverridecommandlockouts        
\usepackage[utf8]{inputenc}
\usepackage[T1]{fontenc}
\usepackage{booktabs}
\usepackage{multirow}
\usepackage{cite}
\usepackage{amsmath,amssymb,amsfonts}
\usepackage{algorithmic}
\usepackage{graphicx}
\usepackage{textcomp}
\usepackage{balance}
\usepackage{cuted}
\usepackage{svg}
\usepackage{booktabs}
\usepackage[normalem]{ulem}
\usepackage{float}
\usepackage{microtype}
\microtypecontext{expansion=basictext,spacing=nonfrench}
\UseMicrotypeSet[protrusion]{basicmath}
\usepackage{caption}
\usepackage{mathtools, amsmath, amssymb}
\usepackage{subcaption}
\usepackage{threeparttable}

\usepackage{tikz}
\usetikzlibrary{spy}

\usepackage{pifont}
\newcommand{\cmark}{\ding{51}}   
\newcommand{\xmark}{\ding{55}}   

\usepackage[table]{xcolor}
\definecolor{purple}{RGB}{160, 110, 210} 
\definecolor{best}{HTML}{8ac9ed}
\definecolor{second}{HTML}{cdc0ed}

\allowdisplaybreaks           

\usepackage{caption}
\newcommand{\etal}{\textit{et al.}}

\usepackage{array}

\usepackage{hyperref}
\hypersetup{
    colorlinks=true,
    linkcolor=red,
    filecolor=magenta,      
    urlcolor=blue,
    citecolor=blue
}

\author{Namritha Lasyapriya Maddali$^{1}$, Rajini Makam$^{2}$, Suresh Sundaram$^{2}$, and Narasimhan Sundararajan$^{3}$
\thanks{$^{1}$Namritha is an Intern at Department of Aerospace Engineering, Indian Institute of Science and Student at Department of Computer Science, PES University,  Bangalore, India.
        {\tt\small namritha.maddali@gmail.com}}%
\thanks{$^{2}$ Department of Aerospace Engineering, Indian Institute of Science,
        Bangalore, India.
        {\tt\small \{rajinimakam, vssuresh\}@iisc.ac.in}}%
        \thanks{$^{3}$ Retired Professor, Nanyang Technological University, Singapore. {\tt\small \{ensundara@ntu.edu.sg\}}}%
}

\begin{document}

\title{$\pi$-SUB: A Physics-Informed Synthetic Underwater Benchmark Dataset for Underwater Image Enhancement} 
\maketitle

\begin{abstract}

This paper presents $\pi$-SUB, a physics-informed framework for generating synthetic underwater benchmark datasets that bridges the synthetic-to-real gap for Underwater Image Enhancement (UIE). The proposed framework extends the classical underwater image formation model by incorporating depth-dependent downwelling irradiance, biologically resolved absorption, and environmental scattering across all ten Jerlov water types, together with independently controllable residual phenomena. Using this framework, the $\pi$-SUB dataset consists of paired synthetic underwater--reference images spanning shallow-to-deep and coastal-to-oceanic environments. Extensive simulation studies have been carried out to evaluate $\pi$-SUB along two criteria namely hyper-realism and generalizability. For \emph{hyper-realism}, $\pi$-SUB attains a global Fr\'{e}chet Inception Distance (FID) that is {46\%} lower than Syrea. For \emph{generalizability}, four state-of-the-art UIE architectures (FUnIE-GAN, Pix2Pix, PUIE-Net, and Phaseformer) are used for comparative evaluation of $\pi$-SUB. These models were independently trained on six datasets including one real and five synthetic datasets and tested on six real-world benchmarks datasets. Across four UIE architectures and six real benchmark datasets, $\pi$-SUB improves UIQM by 4.18\% over PHISWID (next best) and 9.46\% over Syrea (next best), while reducing NIQE by 48.78\% and 23.98\%, respectively. These results establish $\pi$-SUB as a hyper-realistic and generalizable benchmark for developing the next generation of underwater image enhancement methods. \url{https://github.com/airl-iisc/pi-SUB}
\end{abstract}

\begin{IEEEkeywords}
    Physics-Informed, Underwater Image Enhancement, Hyper-realism, Generalization
\end{IEEEkeywords}

\section{Introduction}
\label{sec:intro}

Autonomous underwater vehicles (AUVs) and remotely operated vehicles (ROVs) are widely used for the inspection of subsea communication cables, pipelines, offshore energy infrastructure, and ecological monitoring of coral reefs and marine life~\cite{cong2026comprehensive}. Since sonar provides only limited semantic and textural information, optical imaging remains the primary sensing modality for underwater perception. However, light propagation through water is severely affected by wavelength-dependent absorption and scattering, leading to reduced visibility, color distortion, contrast loss, and veiling light. Underwater image enhancement (UIE) seeks to recover scene radiance by compensating for these degradations and serves as a critical preprocessing step for downstream perception tasks. Supervised UIE, however, requires paired degraded and clean images of the same scene, where the clean reference represents the scene in the absence of the water medium—a condition that is physically impossible to capture~\cite{islam2020fast}. Consequently, bridging the synthetic-to-real domain gap has become essential for developing generalizable UIE models.

Existing paired datasets therefore substitute approximations. Underwater Image Enhancement Benchmark (UIEB)~\cite{li2019underwater} selects, for each image, the visually preferred output of eleven enhancement algorithms; Enhancing Underwater Visual Perception (EUVP)~\cite{islam2020fast} synthesizes a counterpart for each degraded image with a CycleGAN; and Large Scale Underwater Image (LSUI)~\cite{peng2023lsui} extends the same strategy at larger scale. Such approximations are employed because acquiring a distortion-free reference of the same scene without the water medium is physically impossible. The reference images in existing paired datasets exhibit inconsistent colour, contrast, and structural appearance. For example, UIEB and EUVP retain noticeable green colour casts, while LSUI references appear over-bright with a veil-like effect. Consequently, enhancement models tend to learn dataset-specific biases rather than invert the underlying underwater imaging physics. Moreover, the unknown optical conditions of the captured scenes prevent verification against physically valid ground truth, limiting both training and evaluation.

Synthetic data removes this obstacle by constructing the clean reference first and applying a known degradation to it. Two families of methods exist. \emph{Appearance-transfer} methods such as generative translation, physics-guided style transfer, and engine rendering configured from image statistics~\cite{wen2023syreanet, li2025realistic} learn underwater appearance directly from real photographs. They achieve broad visual diversity, but inherit their optical realism from whichever water bodies the reference collection happens to contain and offer no controlled mechanism to extrapolate beyond it. \emph{Physics-based} methods instead derive the degradation from an explicit image formation model, most commonly the Jaffe--McGlamery formulation~\cite{mcglamery1980computer}, and are in principle able to span optical conditions that no photograph in a reference set exhibits. 

Three gaps recur across them. First, they neglect depth-dependent downwelling irradiance by assuming constant surface illumination. Second, they omit biologically driven optical processes, including chlorophyll absorption, CDOM, and chlorophyll-\textit{a} fluorescence. Third, they do not jointly model the volumetric scattering that arises from suspended particulates and distance-dependent haze. Consequently, no existing benchmark simultaneously captures the principal physical and biological processes governing underwater image formation, and the synthetic-to-real gap remains still unresolved.

To close these gaps, this paper proposes a \textbf{P}hysics-\textbf{I}nformed \textbf{S}ynthetic \textbf{U}nderwater \textbf{B}enchmark ($\boldsymbol{\pi}$-SUB) framework. Built on an extended Jaffe--McGlamery model, $\pi$-SUB combines wavelength-dependent inherent optical properties (IOPs) with separate direct-transmission and backscatter attenuation coefficients, models depth-dependent downwelling irradiance across all ten Jerlov water types, resolves absorption into its biological constituents, and augments the formation model with volumetric haze and suspended particulate scattering. Clean references are drawn from photorealistic Unreal Engine renderings together with curated real underwater images with near-zero medium effects, yielding paired synthetic underwater--reference images.

To assess its effectiveness, we introduce a two-axis evaluation protocol based on (i) hyper-realism, which measures the distributional similarity between synthetic and real underwater imagery, and (ii) generalizability, which quantifies the downstream performance of UIE models trained on the benchmark. 
Hyper-realism of $\pi$-SUB has been validated through analyses against real underwater datasets and compared with synthetic datasets SUID~\cite{hou2020suid}, SUIEB~\cite{li2020underwater}, PHISWID~\cite{kaneko2026phiswid} and Syrea~\cite{wen2023syreanet} using Fréchet Inception Distance (FID), out-of-distribution rate, and PCA-based manifold analysis. $\pi$-SUB achieves a 46\% reduction in FID relative to the best existing synthetic benchmark, demonstrating substantially improved distributional alignment with real underwater imagery. To evaluate the generalizability of $\pi$-SUB, four representative UIE architectures (FUnIE-GAN~\cite{islam2020fast}, Pix2Pix~\cite{Isola2017pix2pix}, PUIE-Net~\cite{Fu_2022}, and Phaseformer~\cite{khan2025phase}) are independently retrained on each competing synthetic dataset and the real paired benchmark UIEB using identical hyperparameters, and subsequently evaluated on six real-world benchmarks. Averaged over all evaluations, $\pi$-SUB improves UIQM by 4.2\% over PHISWID (next best) and reduces NIQE by 23.9\% over Syrea (next best). Furthermore, they preserve ~2\% more salient keypoints for feature matching, benefiting downstream robotic tasks such as visual localization, mapping, and autonomous underwater navigation.

In summary, the main contributions of this paper are:

\begin{itemize}
    \item A physics-informed framework, $\pi$-SUB, has been developed to generate paired underwater benchmark datasets from either a set of clean real images or Unreal Engine rendered images by modeling the following, viz the water type, the camera depth, visibility, chlorophyll concentration, and finally optical attenuation.

    \item A modified Jaffe--McGlamery image formation model is developed that incorporates the depth-dependent downwelling irradiance, Jerlov IOPs, separate direct-transmission and backscatter attenuation coefficients, and finally biologically driven optical effects including chlorophyll absorption, CDOM, and fluorescence.

    \item Extensive performance evaluation of $\pi$-SUB has shown that it produces a hyper-realistic and highly generalizable training data, achieving a 46\% lower FID than the next best synthetic benchmark dataset Syrea and consistently improving UIE performance across all other state-of-the-art architectures and real-world benchmark datasets.
\end{itemize}
 and evaluated on real-world data and a downstream feature-matching task.

The paper is organized as follows. Section~\ref{sec:related} reviews existing (both) real and synthetic underwater datasets and highlights  the uniqueness of $\pi$-SUB among them. Section~\ref{sec:pisub} develops the basic image formation model and the $\pi$-SUB image generation framework. Section~\ref{sec:perf} presents the detailed performance evaluation of $\pi$-SUB dataset both distributionally and through the generalization of models trained using $\pi$-SUB. Further, detailed ablation studies, extended hyper-realism analyses, extended metric analysis and statistical significance tests are provided in the supplementary material. These additional results further confirm the robustness and superior generalizability of the proposed $\pi$-SUB framework. For ease of understanding a table with the full expansion of all acronyms is provided. Section~\ref{sec:con} summarizes the conclusions for this study.  

\section{Review of Earlier Work: Existing Datasets for Underwater Image Enhancement}
\label{sec:related}

This section presents a review of both current real-world and synthetic underwater datasets against three properties relevant to supervised training, viz.,: depth-dependent irradiance modeling, Jerlov-grounded inherent optical property (IOP) parameterization, and finally biological optical effects.

\subsection{Current Real-World Underwater Datasets}
UIEB~\cite{li2019underwater} provides 890 real underwater images whose references are the visually preferred output of eleven enhancement algorithms, chosen by volunteer pairwise comparison. EUVP~\cite{islam2020fast} trains a CycleGAN on unpaired data to synthesize a counterpart for each degraded image, and LSUI~\cite{peng2023lsui} extends this strategy to 4{,}279 pairs with semantic and transmission maps. In all these three datasets, the reference is an enhanced version of the underwater image rather than a clean capture, so it encodes the biases of the algorithms or raters used to construct it; Hou~\etal~\cite{hou2020suid} show that model rankings are inconsistent across these benchmarks as a direct consequence. Unpaired collections such as SQUID~\cite{berman2021squid}, OceanDark~\cite{marques2020}, and Real-world Underwater Enhancement (RUIE)~\cite{liu2020ruie} provide no supervisory target and are usable only for no-reference evaluation. No real-world dataset therefore supplies a physically valid clean reference for the captured scene.

\subsection{Current Synthetic Underwater Datasets}

Current synthetic datasets fall into two families. The \emph{physics-based} family derives the degradation from an explicit image formation model. The UWCNN synthetic dataset  (named as SUIEB)~\cite{li2020underwater} applies a global per-type coefficient to per-pixel depth from indoor NYU scenes, extending the formulation to all ten Jerlov water types but using a single attenuation coefficient uniformly across each image and omitting any vertical-depth irradiance term. SUID~\cite{hou2020suid} applies thirty heuristically tuned degradation effects to thirty ground-truth images, individually and in combination, yielding 900 images that are neither IOP-derived nor large enough for training. Physics-Inspired Synthesized Underwater Image Dataset (PHISWID)~\cite{kaneko2026phiswid} introduces per-pixel scene range from RGB-D imagery together with a physics-based marine snow model, but samples vertical depth, background light $B_c$, and water type from uniform distributions rather than deriving attenuation coefficients analytically from measured IOPs, and fixes downwelling irradiance at the surface. Realistic Synthetic Underwater Image Generation (RSUIGM)~\cite{desai2024rsuigm} implements the dual-path model with separate direct and backscatter coefficients and incorporates vertical irradiance, but does not model the biological optical effects characteristic of coastal waters. 

Common to the family, clean images are typically sourced from terrestrial RGB-D datasets, which limits semantic diversity to land-based scenes rather than marine environments such as coral reefs, shipwrecks, and aquatic ecosystems; attenuation parameters are defined globally, heuristically, or per image rather than from measured Jerlov IOPs; depth-dependent irradiance is absent or only partially modeled; scene range is often conflated with vertical depth; and biological optical effects are ignored. The second family learns underwater appearance directly from data, either generatively (WaterGAN~\cite{li2017watergan}, and diffusion-based translation~\cite{zhang2024atlantis}) or through physics-guided style transfer and engine rendering, as in Syrea~\cite{wen2023syreanet} and MUSE~\cite{li2025realistic}. This family inherits its optical realism from the reference distribution rather than from an explicit image formation model, and therefore provides no controlled mechanism to extrapolate to unseen optical conditions; none of its members employs analytic IOP-grounded parameterization.

A summary comparison of these datasets, evaluated across three primary physical properties—depth-dependent irradiance, Jerlov-grounded IOP parameterization, and biological optical effects—and three auxiliary structural properties—invertibility, range-awareness, and metadata—is presented in Table~\ref{tab:dataset_comparison}. The symbols \cmark, $  \sim  $, and \xmark\ denote full, partial, and absent support, respectively. PHISWID and RSUIGM model range-dependent degradation but omit biological optical effects; SUIEB and SUID rely on simplified attenuation models and exclude depth-dependent irradiance; Syrea estimates optical parameters from real images rather than measured IOPs; and MUSE employs a graphics engine yet derives scene parameters from image statistics. As the comparison clearly shows, no existing synthetic dataset jointly incorporates depth-dependent illumination, Jerlov-grounded optical properties, biological spectral effects, and controlled physical parameterization. This gap motivated the development of the $  \pi  $-SUB framework described next.

\begin{table}[t]
\centering
\caption{Physical and structural properties of synthetic underwater image datasets. VD: Vertical depth; IOP: Jerlov IOPs; Bio: Biological effects; Inv.: Invertible; Meta: Metadata.}
\label{tab:dataset_comparison}

\scriptsize
\setlength{\tabcolsep}{2.8pt}
\renewcommand{\arraystretch}{0.95}

\begin{tabular}{lcccccc}
\toprule
Dataset & VD & IOP & Bio & Inv. & Range & Meta \\
\midrule
SUIEB~\cite{li2020underwater}$^*$       & \xmark & $\sim$ & \xmark & $\sim$ & $\sim$ & \xmark \\
SUID~\cite{hou2020suid}$^*$             & \xmark & \xmark & \xmark & $\sim$ & $\sim$ & \xmark \\
Syrea~\cite{wen2023syreanet}$^*$        & \xmark & \xmark & \xmark & \xmark & \xmark & \xmark \\
PHISWID~\cite{kaneko2026phiswid}$^*$    & \xmark & \xmark & \xmark & $\sim$ & \cmark & $\sim$ \\
RSUIGM~\cite{desai2024rsuigm}$^\dagger$ & \cmark & \cmark & \xmark & \cmark & \cmark & \xmark \\
MUSE~\cite{li2025realistic}$^\dagger$   & \xmark & \xmark & \xmark & \xmark & \cmark & \xmark \\
\bottomrule
\end{tabular}

\footnotesize
\noindent
$^*$ Publicly available synthetic datasets used in the experimental comparison;\\ $^\dagger$ not included in the experimental comparison.

\end{table}

\section{Development of a Novel $\pi$-SUB Dataset Generation Framework}
\label{sec:pisub}

In this section, we describe the  development of $\pi$-SUB, a physics-informed synthetic underwater benchmark data-generation framework that transforms a clean reference image into an underwater image with fully recorded per-sample ground truth. This provides a new benchmark dataset referred to as $\pi$-SUB dataset for realistic assessment in underwater image enhancement studies.

\subsection{The $\pi$-SUB Framework}
\label{sec:framework}

The overall $\pi$-SUB generation framework is shown in Figure~\ref{fig:pisub}. The framework accepts three user-defined inputs: (i) a reference image source, selected from either a simulated or real image pool; (ii) the physical water configuration, specified by the Jerlov water type, camera depth, and the corresponding inherent and apparent optical properties (IOPs/AOPs); and (iii) an Augmented Realism configuration, which determines whether suspended particulate matter, volumetric haze, biological effects, or their combination are applied to further degrade the images. The framework then proceeds through three sequential stages. \emph{Stage~I} estimates a dense scene-range map and converts it to a metric underwater range using the water-type-dependent maximum visibility distance $z_{\max}$. \emph{Stage~II} evaluates the proposed modified underwater image formation model to synthesize the deterministic underwater observation $I_b$. Finally, \emph{Stage~III} uses the selected residual optical phenomena to generate the phenomenon-specific and fully augmented subsets that together constitute the $\pi$-SUB dataset. The remainder of this section describes each stage in detail.

\subsubsection{Reference Image Set}

\begin{figure}[htbp]
    \centering
    \includegraphics[width=0.9\linewidth]{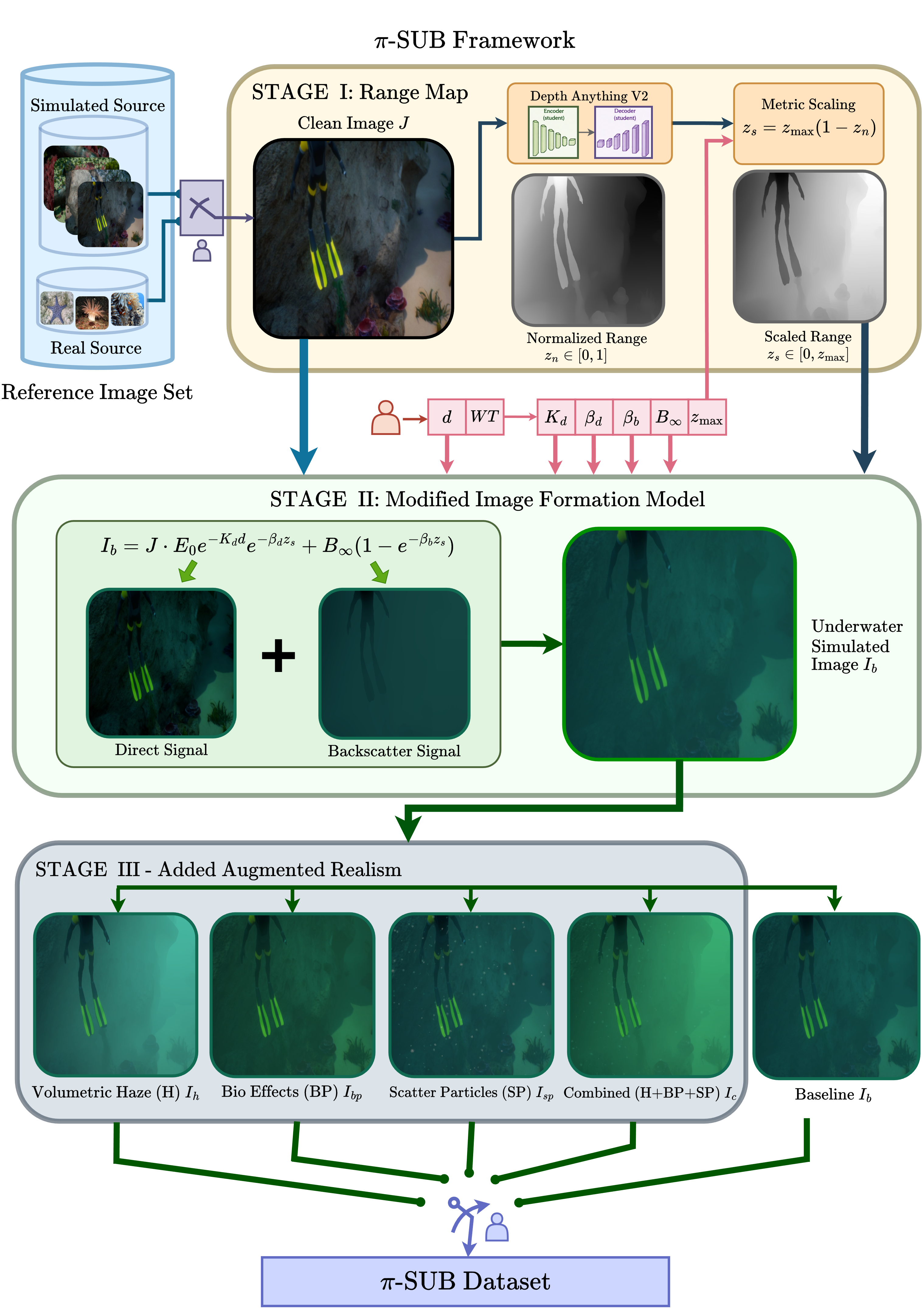}
    \caption{Overall schematic of the \textbf{P}hysics-\textbf{I}nformed \textbf{S}ynthetic \textbf{U}nderwater \textbf{B}enchmark ($\boldsymbol{\pi}$-SUB) dataset generation framework.}
    \label{fig:pisub}
\end{figure}

The leftmost block of Figure~\ref{fig:pisub} represents a clean reference image pool $J$, which is assembled from complementary simulated and real-image sources. The simulated subset is rendered in Unreal Engine without a water medium, such that the recorded radiance corresponds directly to scene reflectance, free from attenuation, scattering, and backscatter, with the sun remaining the sole source of illumination under a downwelling irradiance normalized to $E_0=1$, representing clear, sunny conditions~\cite{akkaynak2018revised}. Every rendered image is accompanied by exact metric range, camera depth, and instance-level semantic annotations. The scenes span representative underwater environments, including coral reefs, ship and aircraft wrecks, rocky seabeds, seagrass, macroalgae, divers, marine animals, and fish schools.

Although Unreal Engine provides physically accurate geometry and controllable ground truth, commercially available asset libraries cannot fully represent the diversity of coral species, marine organisms, and benthic habitats observed in natural underwater ecosystems. To reduce this  gap, a real subset is incorporated by choosing enhanced ground-truth references from UIEB and LSUI. This subset accounts for approximately $10\%$ of the reference pool and is curated rather than sampled: candidates are restricted to shallow-water scenes with minimal residual medium effects. They are retained only where the per-channel RGB response is balanced, so that no residual colour cast is carried into the synthesis. 
Applying the same underwater image formation model to both subsets ensures that the principal difference between them is scene appearance rather than degradation physics, thereby improving the generalization of downstream underwater image enhancement models. 

\subsubsection{Stage~I --- Range Map Estimation} The underwater image formation model requires a per-pixel horizontal range map $z_s$. Exact metric range is available only for the simulated subset through Unreal Engine depth buffers. However, to maintain statistical consistency between simulated and real images, a single monocular depth estimator, Depth Anything V2~\cite{yang2024depth}, is applied uniformly to both subsets. The network output is proximity-like, with pixels closer to the camera driven toward $1$ and distant pixels toward $0$; since the image formation model requires horizontal \emph{range}, this must be inverted before metric scaling. The network predicts a normalized relative depth map $z_n\in[0,1]$, which is converted into metric underwater range through the water-type-dependent calibration $z_s = z_{\max}(1-z_n)$, yielding $z_s\in[0,z_{\max}]$. 
Using the same estimator for both domains prevents a restoration model from exploiting systematic differences in depth statistics. The original Unreal Engine depth buffers are nevertheless retained as reference metadata, enabling quantitative validation of the estimated ranges and providing exact geometry whenever required. Together with the estimated range map, the framework retrieves the assigned physical configuration, namely the camera depth $d$, Jerlov water type, and the derived optical parameters $K_d$, $\beta_d$, $\beta_b$, $B_\infty$, and $z_{\max}$ from the water-type database and passes on to subsequent block.

\subsubsection{Stage~II --- Modified Underwater Image Formation} Given the clean reference image $J$, the estimated range map $z_s$, and the assigned physical configuration, the modified Jaffe--McGlamery image formation model presented in Section~\ref{sec:wco} synthesizes the underwater image $I_b$. The model combines wavelength-dependent direct transmission and backscatter using physically derived optical parameters, producing an analytically invertible degradation whose complete parameter set is recorded for every generated sample. This image $I_b$ is referred to as baseline image and fed as an input to the augmented realism stage.

\subsubsection{Stage~III --- Augmented Realism} The analytical image formation model assumes a homogeneous water column and therefore cannot represent several optical effects commonly observed in natural underwater imagery. Stage~III addresses this limitation by augmenting the deterministic underwater image $I_b$ with three independently controlled phenomena: suspended particulate matter, volumetric multiple scattering (Haze), and biologically induced fluorescence. The mathematical description of these augmented realism is given in Section~\ref{sec:ar}. As illustrated in Figure~\ref{fig:pisub}, each phenomenon is applied individually to generate phenomenon-specific subsets ($I_h$, $I_{sp}$, $I_{bp}$), while an additional subset combines all three effects into $I_c$.

The subsequent sections present the mathematical formulation of Stage I, the physics-based underwater image formation model of Stage II, and the augmented realism model of Stage III.

\subsection{Classical Jaffe--McGlamery Underwater Image Formation Model (IMF)}
\label{sec:ifm}

The physics of underwater light propagation is governed by the Radiative Transfer Equation (RTE), which models light absorption and scattering within a participating medium~\cite{bricaud1998variations}. Under the single-scattering assumption, the RTE reduces to the Jaffe--McGlamery image formation model~\cite{jaffe1990computer}, where the recorded radiance is decomposed into two physically meaningful components: direct transmission from the scene and backscatter generated by the water column.

\begin{equation}
I_b
=
\underbrace{J\,e^{-\beta_d z}}_{D}
+
\underbrace{B_\infty\!\left(1-e^{-\beta_b z}\right)}_{B},
\label{eq:ifm}
\end{equation}

where $J$ is the clean scene radiance, $z$ the horizontal camera-to-scene range, and $\beta_d$, $\beta_b$ the attenuation coefficients of the direct $D$ and backscatter $B$ paths~\cite{akkaynak2018revised}. The two components behave differently with range: direct transmission decays exponentially, whereas backscatter accumulates along the imaging path and asymptotically approaches the veiling light $B_\infty$, so distant regions converge toward a uniform radiance field rather than fading to darkness~\cite{akkaynak2019sea}. They are also governed by different attenuation mechanisms. The direct signal is dominated by absorption, since underwater scattering is strongly forward directed and many scattered photons remain within the camera's acceptance angle, while backscatter is produced by photons redirected toward the sensor and therefore experiences the full beam attenuation rate. Using a single coefficient $\beta = a+b$ for both paths overestimates extinction of the direct signal, so following Akkaynak and Treibitz~\cite{akkaynak2018revised} separate coefficients are adopted,

\begin{equation}
\beta_d(\lambda)
=
a(\lambda)+\delta\,b(\lambda),
\qquad
\beta_b(\lambda)
=
a(\lambda)+b(\lambda)
=
c(\lambda),
\label{eq:betas}
\end{equation}
where $\delta=0.05$ accounts for the limited contribution of forward scattering to direct-path extinction, while the backscatter component follows the full beam attenuation coefficient $c(\lambda)$.

The equation~\eqref{eq:ifm} in this section models radiative transport only along the horizontal imaging path. However, the available illumination reaching the scene depends on the downwelling irradiance, which is attenuated over the vertical water column before interacting with scene surfaces. Both the direct transmission and background veiling light vary with camera depth. Section~\ref{sec:wco} extends the classical formulation by incorporating depth-dependent apparent optical properties and biological effects, leading to the modified underwater image formation model used throughout $\pi$-SUB.

\subsection{Modified Underwater Image Formation (MUIF)}
\label{sec:wco}

In this section, physics and mathematical formulation underlying Stage II of the $\pi$-SUB framework is presented. Building on the image formation model in Section~\ref{sec:ifm}, the direct attenuation ($\beta_d$), backscatter ($\beta_b$ and $B_\infty$), and ambient irradiance are parameterized from the inherent optical properties (IOPs), $(a,b)$, of the water column. The IOPs are intrinsic properties of the medium, whereas the apparent optical properties (AOPs), namely $K_d$, $\beta_d$, $\beta_b$, and $B_\infty$, depend on both the medium and the illumination and viewing geometry. This section derives the AOPs from measured Jerlov IOPs and determines the water-type-specific visibility limits required by the modified underwater image formation (MUIF) model.

The MUIF is parameterized using the ten canonical Jerlov water types, comprising five oceanic (I, IA, IB, II, III) and five coastal (1C, 3C, 5C, 7C, 9C) classes~\cite{jerlov1976marine}. These span the full range of natural optical conditions, from clear oceanic waters to highly turbid coastal environments. The wavelength-dependent absorption $a(\lambda)$ and scattering $b(\lambda)$ are sampled at the representative RGB wavelengths $\lambda=\{650,550,450\}$\,nm following~\cite{jaffe1990computer,akkaynak2018revised}. The IOPs are obtained from the experimentally derived measurements of Solonenko and Mobley~\cite{solonenko2015inherent}. Since measurements are available for six Jerlov types (IB--5C), the remaining types (I, IA, 7C, and 9C) are obtained by monotonic interpolation along the Jerlov clarity scale. Oceanic water types primarily attenuate the red channel, whereas coastal types attenuate all three channels more strongly. Consequently, increasing depth or range progressively reduces scene brightness and contrast.

Although the classical underwater image formation model is parameterized by the inherent optical properties (IOPs) of the water column, it assumes a vertically homogeneous medium and therefore does not distinguish the camera's vertical depth from the horizontal scene range. Consequently, the attenuation of downwelling irradiance $E_d$ with camera depth is neglected, causing the global scene illumination to remain unchanged irrespective of the imaging depth. The proposed MUIF addresses this limitation by explicitly modeling depth-dependent downwelling irradiance as a separate physical process.

\subsubsection{Vertical Irradiance $E_d$}
\label{sec:ed}
The direct transmission component depends not only on the horizontal propagation distance but also on the downwelling irradiance available at the scene. The vertical camera depth must be explicitly modeled, since the available illumination decreases with depth even when the scene range remains unchanged. 

The downwelling irradiance $E_d$ sets the illumination level available at the scene and decays exponentially with depth $d$ at a rate set by the diffuse attenuation coefficient $K_d$. Following Morel and Loisel~\cite{morel1998apparent} and Williamson~\cite{williamson2023depth}, $K_d$ is computed from the IOPs at an optical depth $\zeta = 1$:
\begin{equation}
    K_d(\lambda) = \frac{a(\lambda)}{\mu_d}
    \sqrt{1 + \frac{b(\lambda)}{a(\lambda)}\cdot G(\eta,\mu_d)}
    \label{eq:kd}
\end{equation}
where $\mu_d$ is the mean cosine of the downwelling light just below the water surface ($\mu_d = 0.89$ for oceanic and $0.85$ for coastal waters), and $G$ is the geometry factor:
\begin{equation}
    G(\eta,\mu_d) = \mu_d\!\left(1.3433\,\eta \!-\! 0.01414\right) \!+\! (0.1304 \!+\! 0.272\,\eta)
    \label{eq:gfactor}
\end{equation}
Here $\eta = b_w(\lambda) / b(\lambda)$ is the fraction of total scattering attributable to pure-water (molecular) scattering~\cite{williamson2023depth}, with $b_w(\lambda) = 5.83\times10^{-3}(400/\lambda)^{4.322}$. The parameter $\eta$ captures the transition from particle-dominated scattering ($\eta \rightarrow 0$, turbid coastal water) to molecular scattering ($\eta \rightarrow 1$, clear ocean), and $G$ varies accordingly. The downwelling irradiance at depth $d$ then follows as
\begin{equation}
    E_d(\lambda) = E_0\,e^{-K_d(\lambda) d}
    \label{eq:irradiance}
\end{equation}
with $E_0$ the surface irradiance. Because $K_d < a + b$, light decays more slowly along the vertical than along the horizontal path; depth and range are therefore \emph{not} interchangeable, and each requires its own coefficient.

\subsubsection{Veiling Light $B_\infty$}
The Jerlov IOP tables report the \emph{total} scattering $b(\lambda)$, whereas $B_\infty$ is physically driven by the \emph{backscattering} coefficient $b_b(\lambda) \le b(\lambda)$~\cite{mobley1994light}, since only photons scattered into the rear hemisphere reach the sensor. It is reconstructed from its molecular and particulate contributions:
\begin{equation}
    b_b(\lambda) = \tfrac{1}{2}\,b_w(\lambda) + \tilde{b}_{bp}\,\bigl(b(\lambda) - b_w(\lambda)\bigr),
    \qquad \tilde{b}_{bp} \approx 0.018
    \label{eq:bb}
\end{equation}
where one half of the pure-water scattering $b_w$ is back-directed owing to the near-symmetric molecular phase function, and $\tilde{b}_{bp}\approx 0.018$ is the backscatter fraction of the Petzold average particle phase function, obtained by integrating the measured volume scattering function over the backward hemisphere ($\theta\!\in\![90^\circ,180^\circ]$) as tabulated by Mobley~\cite{mobley1994light}. In the single-scattering regime the veiling light is the backscattered fraction of the depth-attenuated ambient irradiance, saturating at the beam-attenuation rate $c = a+b$:
\begin{equation}
    B_\infty (\lambda)= \frac{b_b(\lambda)\cdot E_d(\lambda)}{a(\lambda) + b(\lambda)}
    \label{eq:binfty}
\end{equation}
Note that $B_\infty$ inherits the depth dependence of $E_d$: the veil dims as the camera descends, exactly as the scene does.

\subsubsection{Biological Constituents of the Medium}
\label{sec:bio}

The preceding formulation models absorption and scattering through bulk optical coefficients. Although sufficient for image formation, these coefficients do not explicitly identify the biological constituents responsible for wavelength-dependent attenuation. In coastal and shallow water the regime represented in the EUVP~\cite{islam2020fast}, LSUI~\cite{peng2023lsui} and RUIE~\cite{liu2020ruie} datasets where the absorption is dominated by an active phytoplankton population that is largely absent from clear oceanic water. Resolving these constituents enables biologically meaningful parameterization while preserving consistency with measured optical properties. First, the chlorophyll concentration is the physical variable that separates the oceanic types from the coastal ones; exposing it converts water type from a categorical label into a continuous, controllable degree of freedom. Second, the same chlorophyll that absorbs light also \emph{re-emits} it, and that emission cannot be represented by any attenuation coefficient.

Three processes contribute. First, selective absorption by chlorophyll-$a$ and its accessory pigments. Second, co-absorption by Chromophoric Dissolved Organic Matter (CDOM), a photo-degradation byproduct of phytoplankton metabolism whose concentration is correlated with that of chlorophyll. Third, inelastic chlorophyll fluorescence, in which absorbed blue and red photons are re-emitted near $685$\,nm. The first two are \emph{elastic} losses and belong in the medium; the third is an \emph{additive emission} called Fluorescence.

Following the experimentally-derived constituent decomposition of Solonenko and Mobley~\cite{solonenko2015inherent}, the measured Jerlov absorption is reconstructed as the sum of pure water, chlorophyll and CDOM:
\begin{equation}
    a(\lambda) = a_w(\lambda) + a_{chl}(\lambda) + a_{cdom}(\lambda)
    \label{eq:a_total}
\end{equation}
where $a_w$ is the pure-water absorption~\cite{buiteveld1994optical}. The chlorophyll term follows the Bricaud~\etal~\cite{bricaud1998variations} power law:
\begin{equation}
    a_{chl}(\lambda) = A_\lambda\,C_{chl}^{E_\lambda}
    \label{eq:achl}
\end{equation}
where $A_\lambda$ and $E_\lambda$ are wavelength-dependent pigment coefficients and $C_{chl}$ is the chlorophyll concentration (mg/m$^3$). CDOM is coupled to chlorophyll rather than treated as an independent constituent, since it is predominantly a photo-oxidation byproduct of phytoplankton material and co-varies with $C_{chl}$~\cite{bricaud1998variations}:
\begin{equation}
    a_{cdom}(\lambda) = a_{chl}(440)\,M\,e^{-\alpha(\lambda - 440)}
    \label{eq:acdom}
\end{equation}
where $M$ (unitless) and $\alpha$ (nm$^{-1}$) are per-type CDOM fitting parameters. The per-type values of $C_{chl}$ and $(M,\alpha)$ are taken from the fits of~\cite{solonenko2015inherent} and details are given in the supplementary material.

The scattering coefficient admits an analogous decomposition into a molecular term and two particulate modes:
\begin{equation}
    b(\lambda) = b_w(\lambda) + \underbrace{B_s\,b_s(\lambda) + B_l\,b_l(\lambda)}_{b_p(\lambda)}
    \label{eq:b_total}
\end{equation}
where $b_w$ is the pure-water (molecular) scattering term introduced in Section~\ref{sec:ed}, and $b_p$ splits into small- and large-particle modes $b_s(\lambda)=1.1513(400/\lambda)^{1.7}$ and $b_l(\lambda)=0.3411(400/\lambda)^{0.3}$, with per-type concentrations $(B_s,B_l)$ from the same Solonenko--Mobley fit~\cite{solonenko2015inherent}.
Absorption is thus biologically resolved while scattering enters through bulk particulate modes, but both are drawn from a single measured constituent basis, so $a$ and $b$ remain mutually consistent. Because $C_{chl}$ now appears explicitly, it can be varied independently of the nominal water type, allowing a fixed scene geometry to be rendered across a productivity gradient.

Having resolved the absorption and scattering coefficients into their physical constituents, the complete set of inherent optical properties is now available. These properties not only parameterize the image formation model but also determine the maximum distance over which scene contrast can be preserved. The corresponding visibility limits are therefore derived next and used to bound the scene range during dataset synthesis.

\subsubsection{Visibility Limits}
The backscatter term saturates at a range beyond which no scene contrast survives, and this defines the physical extent of the scene rather than an arbitrary rendering choice. Koschmieder's Law~\cite{koschmieder1925theorie} gives the extinction of visible contrast along the line of sight as $C_z = C_0 e^{-\beta z}$. The beam contrast analysis of Lee \etal~\cite{lee2016visibility} reports a detection threshold of $\varepsilon \approx 0.05$ for the human eye and $\varepsilon \approx 0.01$ for camera systems~\cite{bricaud1998variations}. Since $\pi$-SUB models images captured by a camera, the camera threshold $C_z/C_0 = \varepsilon \approx 0.01$ is used throughout. The maximum scene range is therefore
\begin{equation}
    z_{\max} = \frac{-\ln\varepsilon}{\,\beta_{\min}\,}, \qquad \beta_{\min} = \min_\lambda\bigl(a(\lambda) + b(\lambda)\bigr)
    \label{eq:zmax}
\end{equation}
where $\beta_{\min}$ is the minimum beam attenuation across the three channels, since visibility is governed by the channel that penetrates furthest. The corresponding vertical limit $d_{\max}$ follows from $K_d$. Horizontal visibility $z_{\max}$ exceeds $200$\,m in clear oceanic water (Type~I) and collapses below two metres in turbid Type~9C coastal water, while $d_{\max}$ remains consistently larger because $K_d < a+b$. Both bounds are strongly type-dependent, and $\pi$-SUB treats them as such rather than fixing a single scene extent across all water types; the complete per-type $(z_{\max}, d_{\max})$ values are tabulated in the supplementary material.

Combining the separated attenuation coefficients, depth-dependent irradiance, biologically resolved absorption, and veiling light with the classical Jaffe--McGlamery formulation yields the image formation model adopted by $\pi$-SUB:
\begin{equation}
I_b(\lambda)
=
\underbrace{J\,E_d(\lambda)e^{-\beta_d(\lambda)z_s}}_{\text{direct}}
+
\underbrace{B_\infty(\lambda)\!\left(1-e^{-\beta_b(\lambda)z_s}\right)}_{\text{backscatter}},
\label{eq:ib}
\end{equation}
where $I_b$ is referred as Baseline image, $E_d=E_0e^{-K_dd}$, $\beta_d=a+\delta b$, $\beta_b=a+b$, $B_\infty=b_bE_d/(a+b)$, and $a=a_w+a_{chl}+a_{cdom}$. Unlike the conventional Jaffe--McGlamery model, the proposed formulation explicitly separates direct and backscatter attenuation, incorporates depth-dependent illumination through $E_d$, and resolves absorption into measured biological constituents while remaining deterministic and analytically invertible.

For each clean reference image, $\pi$-SUB evaluates \eqref{eq:ib} in linear radiometric space using the assigned Jerlov water type, vertical camera depth $d$, surface irradiance $E_0$, and per-pixel scene range $z_s$. Every image is synthesized under all ten Jerlov water types and multiple camera depths, with $z_s$ obtained by calibrating the normalized Depth Anything V2 prediction $z_n$ to the water-type-specific visibility limit,
\begin{equation}
z_s=z_{\max}(1-z_n),
\label{eq:zs}
\end{equation}
where the same calibration is applied to both simulated and real image pools. Unlike previous formulations that conflate camera depth and scene range, $\pi$-SUB models vertical depth through the global illumination term $E_d$ and horizontal range through the spatially varying attenuation and backscatter terms.

\begin{figure}[htbp]
    \centering
    \includegraphics[width=0.7\linewidth]{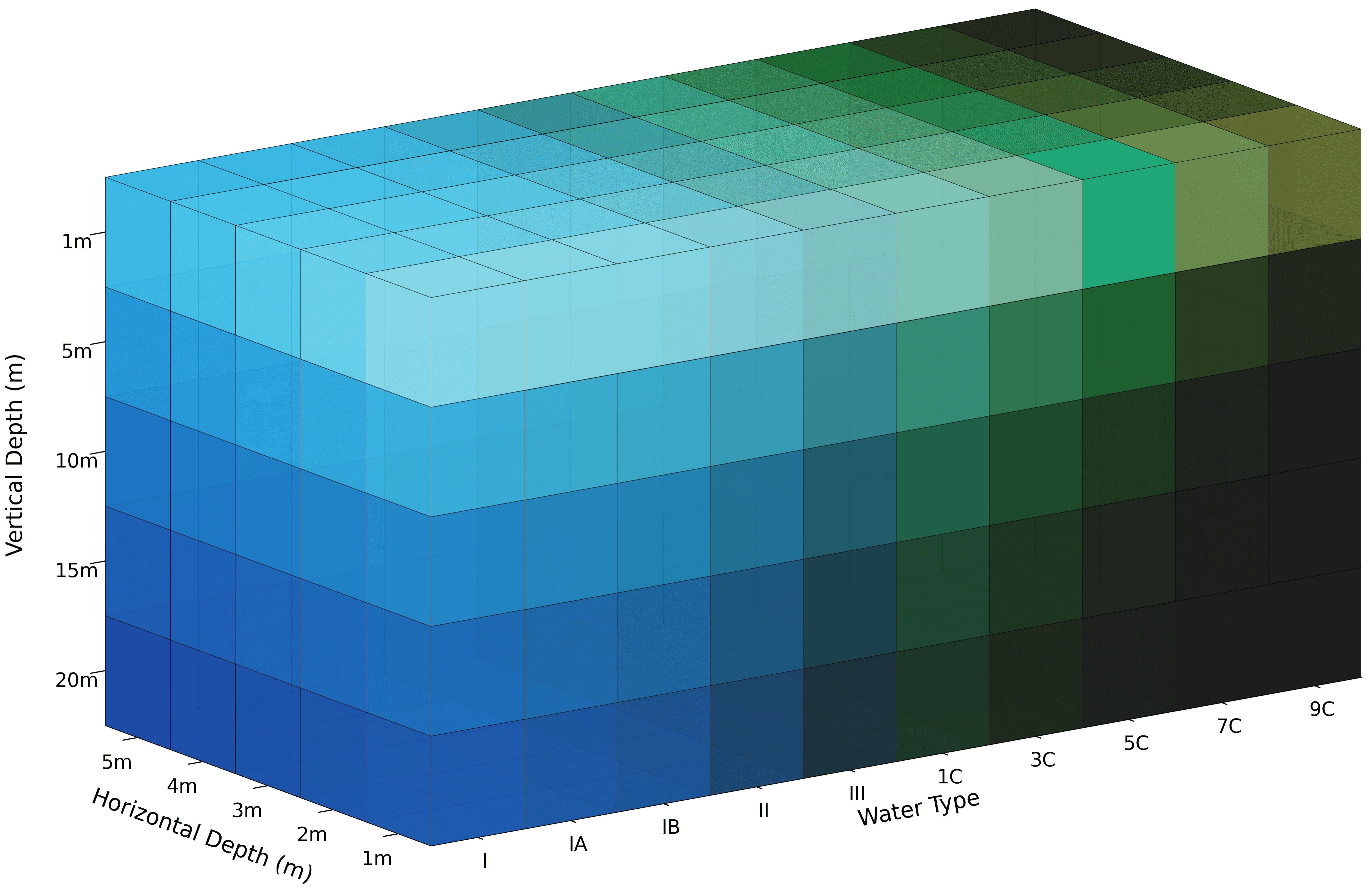}
    \caption{Joint degradation over Jerlov type, range, and depth: oceanic types (I--III) retain a blue--green cast, coastal types (1C--9C) darken uniformly.}
    \label{fig:voxel}
\end{figure}

Fig.~\ref{fig:voxel} visualizes the resulting joint degradation surface produced by evaluating Eq.~\eqref{eq:ib} across all ten Jerlov types, horizontal range, and vertical depth. Oceanic types (I--III) retain a blue-green cast since only the red channel is stripped, while coastal types (1C--9C) darken across all channels. Along either axis, intensity falls monotonically as the corresponding attenuation term decays, confirming that range and depth are governed by distinct, non-interchangeable coefficients.

The modified image formation model accurately reproduces the dominant deterministic effects of underwater light propagation. However, several residual phenomena observed in real underwater imagery arise from complex volumetric interactions and stochastic environmental variability that are not captured by analytical radiative transport models. The following section addresses these effects through the proposed Augmented Realism stage.

\subsection{Augmented Realism}
\label{sec:ar}

Having established the deterministic image formation model in Stage II, this section describes Stage III of the $\pi$-SUB framework, which augments $I_b$ with residual optical phenomena that are not captured by the analytical model. The proposed Augmented Realism overlays three independently controlled residual phenomena on the deterministic underwater image $I_b$: suspended particulate matter, volumetric haze, and biological effects. Each phenomenon is generated as a separate subset, enabling controlled evaluation of individual optical degradations as well as their combined effects.

\subsubsection{Suspended Particulate (SP) Matter}
Suspended particles are composited as a procedural layer on top of $I_b$ following recent underwater rendering practice~\cite{kaneko2023marine, bagoren2026surfslam}. Each particle is rendered as an oriented anisotropic Gaussian whose opacity $\alpha_i$ is modulated by an occlusion term derived from the normalized range map, so that particles lying behind scene geometry are suppressed. The particulate layer is composited as
\begin{equation}
    I_{sp}
    =
    I_b
    +
    \lambda\sum_{i=1}^{N}\alpha_i\mathbf{c}_i,
    \label{eq:particle_composite}
\end{equation}
where $N$ is the particle count, $\mathbf{c}_i$ the particle colour, and $\lambda$ the overall intensity. The full parameterization of $\alpha_i$, including the size, elongation, and occlusion terms, is given in the supplementary material.

\subsubsection{Volumetric Haze}
The single-scattering term in \eqref{eq:ib} underestimates the diffuse veil produced by multiple scattering. $\pi$-SUB approximates this effect with a continuous range-dependent haze~\cite{mobley1994light, jaffe1990computer}:
\begin{equation}
    I_h=I_b(1-\rho)+\mathbf{c}_{\text{haze}}\rho,\qquad
    \rho=(1-z_n)^{1/s_h},
    \label{eq:haze}
\end{equation}
where $z_n$ is the normalized scene range, $\mathbf{c}_{\text{haze}}$ is the spectrally tinted haze colour, and $s_h$ controls the severity.

\subsubsection{Biological Effects (BE)}
Chlorophyll and CDOM absorption are already encoded in the base model through $a$ in Section~\ref{sec:bio}. The remaining biological phenomenon is inelastic fluorescence, which introduces additive radiance and therefore cannot be represented by attenuation alone~\cite{gordon1979diffuse, zhai2018radiative}. The fluorescence yield is not constant with depth. Near the surface, high photosynthetically active radiation (PAR) drives photo-inhibition, which dissipates absorbed energy as heat and suppresses the quantum yield. As the scalar irradiance attenuates with depth, photo-inhibition relaxes, the yield rises to a characteristic subsurface maximum, and the fluorescence signal peaks before photon starvation at depth extinguishes it~\cite{maritorena2000influence}. This non-monotonic depth profile distinguishes fluorescence from every other modeled phenomenon, all of which decrease monotonically with depth. The resulting RGB fluorescence source term is
\begin{equation}
    \mathbf{F}_{em} = \phi_C\cdot a_\phi\cdot\mathbf{f}_{em},
    \qquad \mathbf{f}_{em} = [0.97,\;0.03\;,0.00]^\top
    \label{eq:fem}
\end{equation}
where $\phi_C$ is the depth-resolved quantum yield obtained from the photo-inhibition model of Zhai~\etal~\cite{zhai2018radiative}, $a_\phi$ is the phytoplankton absorption evaluated at the two chlorophyll excitation peaks following the Bricaud~\etal{} parameterization~\cite{bricaud1998variations}, and $\mathbf{f}_{em}$ maps the $685$\,nm emission band onto the RGB channels. The complete four-stage derivation, all photo-physical constants, and the green-normalized chromatic refinement applied to productive coastal waters~\cite{kirk1994light,mobley1994light} are given in the supplementary material. Fluorescence and chromatic refinement become increasingly pronounced with chlorophyll concentration, producing negligible changes for clear oceanic waters (Types~I--IB) and stronger coastal coloration for productive Jerlov types (1C--9C). The proposed fluorescence model is physically motivated and consistent with established radiative transfer principles; it has not yet been independently validated against \emph{in-situ} measurements of chlorophyll fluorescence and absorption, and is identified as future work.

Together with the suspended particulate matter and volumetric haze models presented previously, the biological effects complete the proposed Augmented Realism stage. The independently controlled combinations of these residual phenomena define the final organization of the $\pi$-SUB benchmark.

The $\pi$-SUB comprises a deterministic MUIF subset ($I_b$) together with phenomenon-specific variants: MUIF+Haze ($I_h$), MUIF+BE ($I_{bp}$), and MUIF+SP ($I_{sp}$), as well as a fully augmented subset, MUIF+Haze+BE+SP ($I_c$), which combines all three residual phenomena. This organization enables controlled evaluation of individual optical phenomena across Jerlov water types while also providing a hyper-realistic composite benchmark that more closely reflects natural underwater imaging conditions. Each sample is accompanied by complete metadata, including the Jerlov water type, camera depth $d$, chlorophyll concentration $C_{chl}$, visibility limits $(z_{\max}, d_{\max})$, surface irradiance $E_0$, and channel-wise optical parameters, ensuring full reproducibility of the synthesis process. By combining physically grounded modified image formation with independently controllable residual effects, $\pi$-SUB spans a broad distribution of underwater optical conditions, reducing the simulation-to-reality gap and improving the generalizability of underwater image enhancement models trained on the benchmark.

The released benchmark is generated from $N=117$ clean reference images drawn from the simulated and real pools of Section~\ref{sec:framework}. Each reference is synthesized under all ten Jerlov water types for the $I_b$, $I_{sp}$, and $I_{bp}$ subsets, under ten water types at three severity settings for $I_h$ and $I_c$, and under thirty randomized effect--depth--water-type combinations for the held-out test split, giving $117\times120=14{,}040$ paired synthetic underwater--reference images in total. The per-subset breakdown and the metadata schema are tabulated in the supplementary material. 

\section{Performance Evaluation of $\pi$-SUB} \label{sec:perf}
In the preceding section, the mathematical foundations of $\pi$-SUB, introducing a physics-informed synthesis framework that combines measured Jerlov optical properties, depth-dependent irradiance, and biologically driven spectral modulation to generate paired underwater images was presented. In this section, validation of the $\pi$-SUB against its two major design objectives: (i) reproducing the distribution of real underwater imagery (\emph{hyper-realism}), and (ii) to show how good the generalization of underwater image enhancement UIE models trained using the proposed dataset is achieved (\emph{generalizability}).

Hyper-realism is assessed by comparing the feature-space distribution of $\pi$-SUB against SUID~\cite{hou2020suid}, SUIEB~\cite{li2020underwater}, Syrea~\cite{wen2023syreanet}, and PHISWID~\cite{kaneko2026phiswid}, using the combined distribution of five real benchmarks (UIEB~\cite{li2019underwater}, RUIE~\cite{liu2020ruie}, SQUID~\cite{berman2021squid}, FishTrac~\cite{dawkins2024fishtrack23}, OceanDark~\cite{marques2020}) as reference, quantified through Fr\'{e}chet Inception Distance (FID)~\cite{heusel2017FID}, out-of-distribution (OOD) rate, and principal component analysis. Generalizability is assessed by independently training four UIE architectures --- Pix2Pix~\cite{Isola2017pix2pix}, FUnIE-GAN~\cite{islam2020fast}, Phaseformer~\cite{khan2025phase}, and PUIE-Net~\cite{Fu_2022} --- on each candidate training dataset and evaluating them on six real-world benchmarks, which isolates the influence of the training data from that of the architecture.

\subsection{Hyper-realism Validation}
\label{sec:dataset_validation}

Each image is first mapped to a physically interpretable perceptual feature space, on which four complementary analyses are performed: perceptual clustering, global and cluster-wise FID, the out-of-distribution rate, and a PCA projection of the combined distributions.

\subsubsection{\textbf{Perceptual Feature Clustering}}

\begin{figure}[!t]
\centering

\includegraphics[width=\columnwidth]{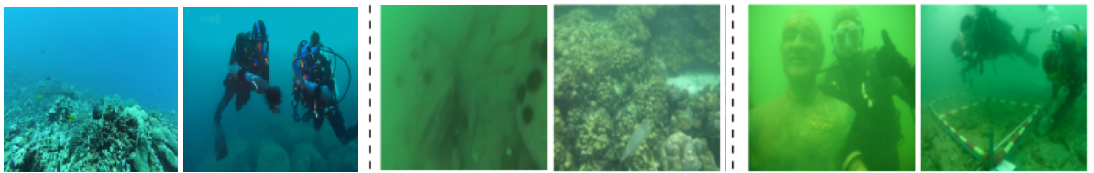}

\vspace{1mm}

\includegraphics[width=\columnwidth]{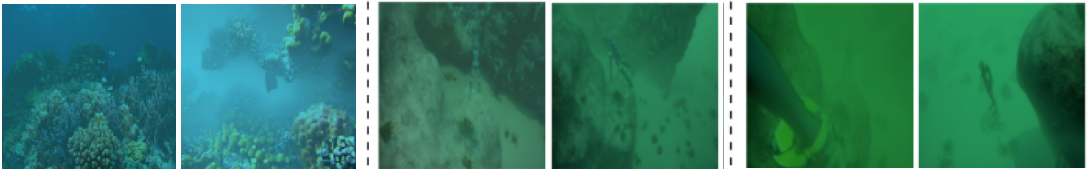}

\caption{Representative images from three perceptual clusters obtained using $K{=}7$ K-means on the combined real feature space (UIEB, RUIE, SQUID, FishTrac, and OceanDark). (a) Real underwater images. (b) Corresponding $\pi$-SUB images.}
\label{fig:cluster_grids}

\end{figure}

Each image is first represented by a physically interpretable perceptual feature vector comprising the mean and standard deviation of brightness ($L^*$) and chroma, together with the mean dark-channel value, the mean HSV saturation, and the chromatic coordinates ($a^*$, $b^*$). These features are directly related to the optical effects of absorption, scattering, and illumination, unlike generic deep features that lack physical interpretability. A $K$-means model ($K=7$) is fitted to the standardized feature vectors extracted from the combined real datasets (UIEB, RUIE, SQUID, FishTrac, and OceanDark), partitioning the real data into seven perceptually distinct clusters. $\pi$-SUB is the only synthetic dataset represented in all seven clusters, indicating broad coverage of the perceptual characteristics observed in real underwater imagery. Figure~\ref{fig:cluster_grids} shows a sample cluster containing a real underwater image and the corresponding $\pi$-SUB-generated images. The generated images are visually close to the real images. In contrast, the other synthetic datasets exhibit incomplete coverage: SUIEB is not represented in cluster C4, PHISWID is not represented in cluster C6, and the remaining synthetic datasets show similar gaps. Representative images from each cluster are provided in the supplementary material, confirming that $\pi$-SUB consistently matches the color, contrast, and haze characteristics of the corresponding real images, whereas competing datasets either have sparse representation or exhibit noticeable visual deviations in several clusters.

\subsubsection{\textbf{Global and cluster-wise FID}}

FID measures distributional similarity by passing each set through an Inception-v3 network and computing the Fr\'{e}chet distance between the resulting feature distributions; lower is better. It is computed both globally and per perceptual cluster, so that the comparison reflects performance within each physically grounded optical regime rather than only in aggregate.

The cluster-wise and global FID scores of each synthetic dataset against the combined real benchmark pool are reported in Table~\ref{tab:fid_clusters}. $\pi$-SUB achieves the lowest global FID at 95. This compares to 176 for Syrea, 192 for SUIEB, 227 for PHISWID, and 230 for SUID. This result holds across every cluster individually. SUIEB does not populate cluster C4, PHISWID does not populate cluster C6, and the remaining datasets, including PHISWID, score roughly 50 to 200 FID points above $\pi$-SUB in every cluster where they are populated. The cluster-wise breakdown shows that $\pi$-SUB's lower distance to the real distribution is consistent across the full range of perceptual conditions present in real underwater imagery, and not just on average.

\renewcommand{\thefootnote}{\fnsymbol{footnote}}
\begin{table}[htbp]
    \centering
    \caption{Cluster-wise and global FID against real distributions (UIEB + RUIE + SQUID + OceanDark + FishTrac).
    Lower is better. Best per column in \cellcolor{best}{cyan}.}
    \setlength{\tabcolsep}{4.5pt}
    \begin{tabular}{lccccccccc}
        \toprule
        Dataset & C1 & C2 & C3 & C4 & C5 & C6 & C7 & Global \\
        \midrule
        SUID & 278 & 348 & 258 & 207 & 274 & 355 & 258 & 230 \\
        SUIEB & 254 & 248 & 316 & --- & 290 & 394 & 215 & 192 \\
        PHISWID & 291 & 271 & 278 & 256 & 392 & --- & 272 & 227 \\
        Syrea &\cellcolor{second}{249} &\cellcolor{second}{235} &\cellcolor{second}{232} & \cellcolor{second}{203} &\cellcolor{second}{234} &\cellcolor{second}{262} &\cellcolor{second}{225} &\cellcolor{second}{176} \\

        \textbf{$\pi$-SUB} &\cellcolor{best}{\textbf{192}} &\cellcolor{best}{\textbf{196}} &\cellcolor{best}{\textbf{126}} &\cellcolor{best}{\textbf{105}} &\cellcolor{best}{\textbf{183}} &\cellcolor{best}{\textbf{232}} &\cellcolor{best}{\textbf{142}} &\cellcolor{best}{\textbf{95}}\\
        \bottomrule
    \end{tabular}  \label{tab:fid_clusters}
\end{table}

\subsubsection{\textbf{Out-of-distribution (OOD) rate}}
\label{sec:ood}

FID does not indicate what fraction of synthetic images lie entirely outside the real manifold. An image is flagged as out-of-distribution (OOD) if its distance to the nearest real-data cluster centroid, in the same perceptual feature space, exceeds the 95th percentile of distances observed among real images.

Only $2.00\%$ of the 14{,}040 $\pi$-SUB images are flagged as OOD, against $4.02\%$ for PHISWID (9{,}426 images), $12.23\%$ for Syrea (21{,}826), $19.26\%$ for SUIEB (11{,}592), and $27.33\%$ for SUID (900). This reduction reflects the benefit of grounding optical parameters in measured Jerlov IOPs and modelling depth-dependent irradiance and biological spectral effects rather than approximating them heuristically. The $\pi$-SUB images that are flagged arise predominantly from deep coastal water with strong biological absorption and no artificial illumination: physically valid conditions that are structurally absent from real benchmark collections, since field deployments rarely operate under them. The OOD fraction therefore reflects coverage beyond the reach of current real benchmarks rather than a mismatch with underwater optics.

\subsubsection{\textbf{Principal Component Analysis (PCA)}}
Projecting the perceptual feature space onto its first two principal components, which explain 40.2\% and 22.2\% of the total variance, provides a visual comparison of the dataset distributions. The convex hull of $\pi$-SUB closely overlaps that of the real data, with nearby centroids, whereas Syrea occupies a partially overlapping but shifted region. SUIEB and PHISWID cluster together in an offset region along PC1, reflecting their characteristic brightness and chromatic differences, while SUID occupies only a small region of the feature space because of its limited size. The corresponding PCA projection is provided in the supplementary material.

Further checks are reported in the supplementary material and support the same conclusion. Cluster-wise visual inspection confirms that $\pi$-SUB images assigned to each real cluster carry the correct water type and depth label, a per-image nearest-neighbour analysis shows that $90\%$ of real images find a $\pi$-SUB counterpart at a smaller feature distance than for any competing dataset, and cross-validation against four SQUID survey sites recovers mean matched depths that fall within each site's reported range.

Together these analyses establish the hyper-realism of $\pi$-SUB: the lowest FID to the real distribution both globally (95 against 176 for the next-best dataset) and in every perceptual cluster, an OOD rate of $2.00\%$, the greatest overlap with real data in PCA space, and the only synthetic dataset covering all seven perceptual regimes. The ranking is not an artifact of the clustering configuration: repeating the analysis with eight clusters over a partially overlapping benchmark pool leaves $\pi$-SUB lowest overall and in every populated cluster, as reported in the supplementary material. Distributional similarity alone does not guarantee learning effectiveness, so the next subsection evaluates whether it translates into better generalization.

\subsection{Generalization Performance Evaluation}
\label{sec:benchmarking}

Generalization is measured through the enhancement quality of models trained on each candidate dataset. Four architectures spanning distinct inductive biases are used: Pix2Pix~\cite{Isola2017pix2pix}, a conditional adversarial baseline; FUnIE-GAN~\cite{islam2020fast}, a generative adversarial network; PUIE-Net~\cite{Fu_2022}, a probabilistic CNN; and Phaseformer~\cite{khan2025phase}, a frequency-domain vision transformer. Each is trained independently on the real paired dataset UIEB~\cite{li2019underwater} and on the synthetic datasets SUID~\cite{hou2020suid}, SUIEB~\cite{li2020underwater}, PHISWID~\cite{kaneko2026phiswid}, Syrea~\cite{wen2023syreanet}, and $\pi$-SUB, giving twenty-four trained models. All models are retrained from scratch under identical hyperparameters, input resolution, and schedule, with the number of training pairs equalized across datasets; the complete protocol, including the subsampling used to equalize dataset scale, is given in the supplementary material.

These twenty-four models are evaluated on six real benchmark underwater datasets, each contributing a distinct visual and optical character to the evaluation suite. 
UIEB~\cite{li2019underwater} is the largest and most heterogeneous of these, comprising a broad collection of shallow-water scenes that include divers, fish, coral, rocks, and underwater cables and structures, captured under a wide range of illumination and turbidity conditions. U45 is a designated subset of UIEB, isolating forty-five of its most severely degraded images, characterized by low contrast, non-uniform lighting, and pronounced color casts, and is used to probe enhancement performance under the hardest conditions the parent dataset offers.
RUIE~\cite{liu2020ruie} spans a wide variety of blue- and green-tinted underwater scenarios collected for underwater robotic perception tasks, and its imagery was later used as the source material from which LSUI~\cite{peng2023lsui} generated its enhanced model-based references. 
OceanDark~\cite{marques2020} consists of low-light scenes captured at deep-sea sites under artificial illumination, which impart a characteristic green-dominant cast. SQUID~\cite{berman2021squid} and FishTrac~\cite{dawkins2024fishtrack23}, by contrast, are both drawn from open, relatively clear oceanic waters and exhibit a predominantly blue color cast, with SQUID additionally providing stereo image pairs of coral reef scenes and FishTrac capturing fish-tracking footage in similarly blue-dominant conditions.

\subsubsection{\textbf{Evaluation Metrics}}
Because clean references are unavailable for real underwater imagery, enhancement is scored with no-reference metrics. The Underwater Image Quality Measure (UIQM)~\cite{panetta2018} is a weighted sum of colorfulness, sharpness, and contrast terms,
\begin{equation}
\text{UIQM} = c_1 \times \text{UICM} + c_2 \times \text{UISM} + c_3 \times \text{UIConM}, \label{uiqm}
\end{equation}
where UICM, UISM, and UIConM quantify colour quality, edge sharpness, and local contrast, with empirically determined weights $c_1=0.0282$, $c_2=0.2953$, $c_3=3.5753$~\cite{panetta2018}; higher is better. The Natural Image Quality Evaluator (NIQE)~\cite{mittal2013} measures the deviation of patch statistics from a model of natural scene statistics learned from undistorted images, requiring neither reference images nor human opinion scores; lower values indicate outputs that are statistically closer to natural imagery. A secondary colour-based metric, UCIQE~\cite{Yang2015}, is reported in the supplementary material.

\subsubsection{Qualitative Analysis}
\begin{figure*}[!t]
\centering
\def\img#1{%
\raisebox{-.5\height}{%
\includegraphics[
width=0.068\textwidth,
height=0.068\textwidth]{#1}}}
\def\hdr#1{\makebox[0.068\textwidth][c]{\scriptsize #1}}
\def\lbl#1{\rotatebox[origin=c]{90}{\scriptsize #1}}

\setlength{\tabcolsep}{0pt}
\renewcommand{\arraystretch}{1.0}

\begin{tabular}{
c@{\hspace{2pt}}
cccc@{\hspace{3pt}}
cccc@{\hspace{3pt}}
cccc
}

& \multicolumn{4}{c}{\textbf{UIEB}} &
\multicolumn{4}{c}{\textbf{OceanDark}} &
\multicolumn{4}{c}{\textbf{SQUID}} \\

& \hdr{(a)} & \hdr{(b)} & \hdr{(c)} & \hdr{(d)}
& \hdr{(a)} & \hdr{(b)} & \hdr{(c)} & \hdr{(d)}
& \hdr{(a)} & \hdr{(b)} & \hdr{(c)} & \hdr{(d)} \\

\lbl{Raw}
& \img{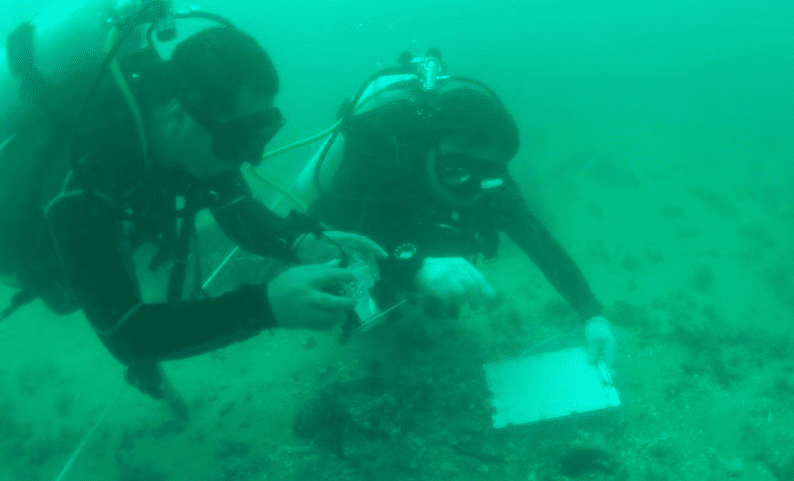}
& \img{set/inference/raw/raw_56.png}
& \img{set/inference/raw/raw_56.png}
& \img{set/inference/raw/raw_56.png}

& \img{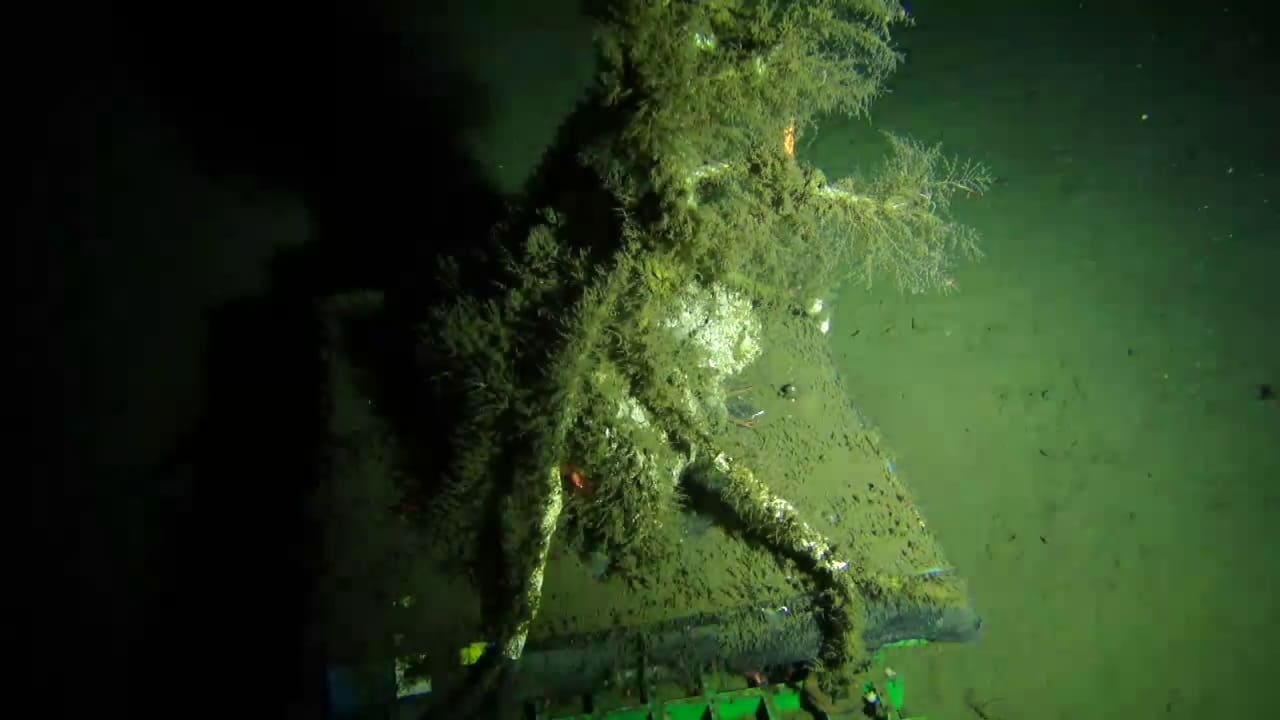}
& \img{set/inference/raw/35.jpg}
& \img{set/inference/raw/35.jpg}
& \img{set/inference/raw/35.jpg}

& \img{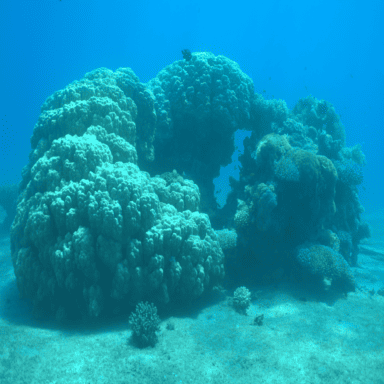}
& \img{set/inference/raw/TRAIN_INPUT_RGT_4495.png}
& \img{set/inference/raw/TRAIN_INPUT_RGT_4495.png}
& \img{set/inference/raw/TRAIN_INPUT_RGT_4495.png}
\\
\lbl{UIEB}
& \img{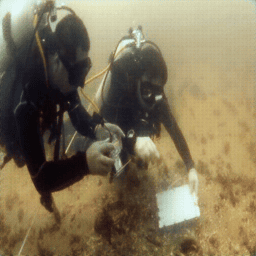}
& \img{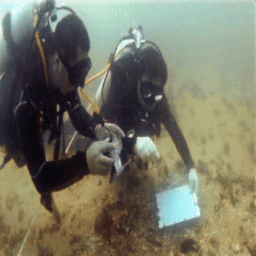}
& \img{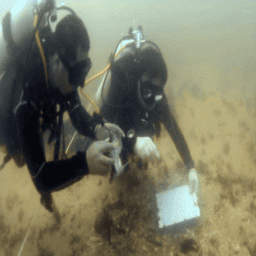}
& \img{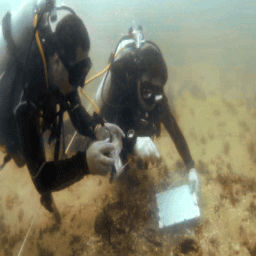}

& \img{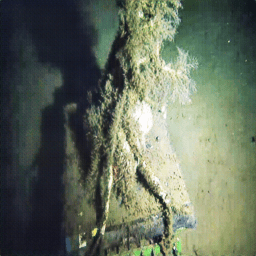}
& \img{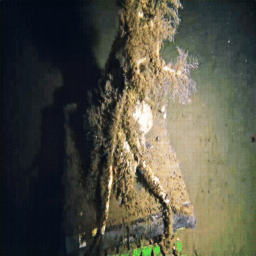}
& \img{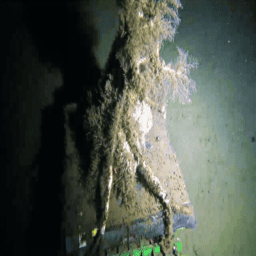}
& \img{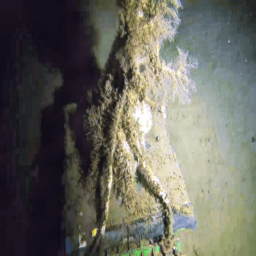}

& \img{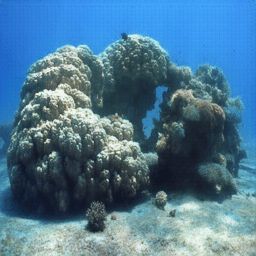}
& \img{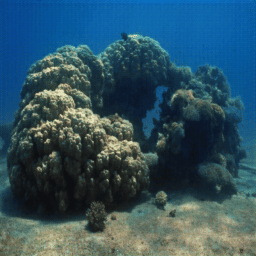}
& \img{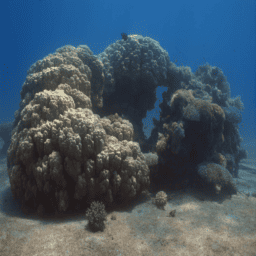}
& \img{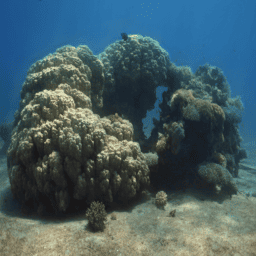}
\\

\lbl{SUID}
& \img{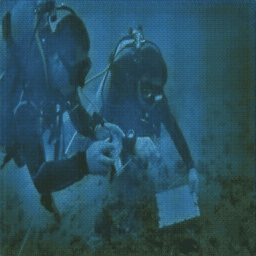}
& \img{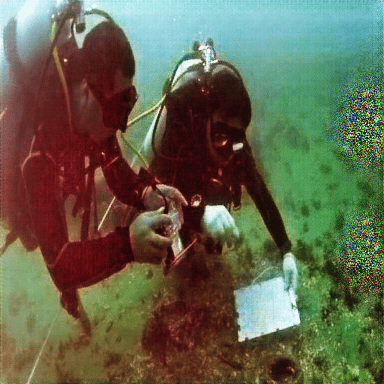}
& \img{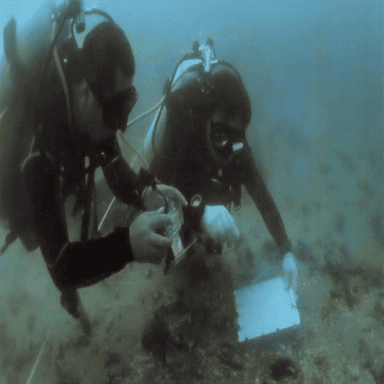}
& \img{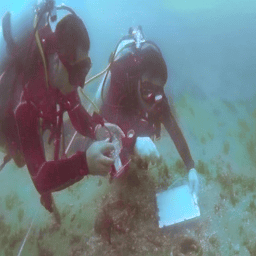}

& \img{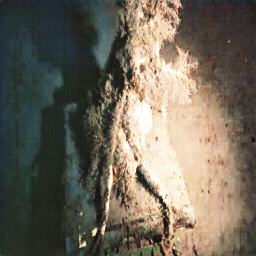}
& \img{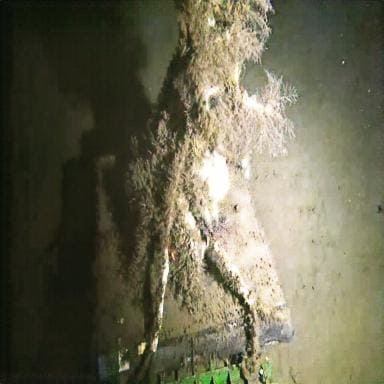}
& \img{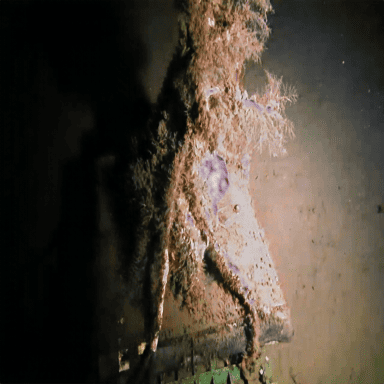}
& \img{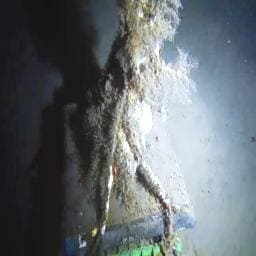}

& \img{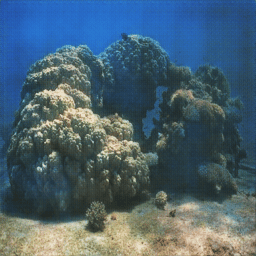}
& \img{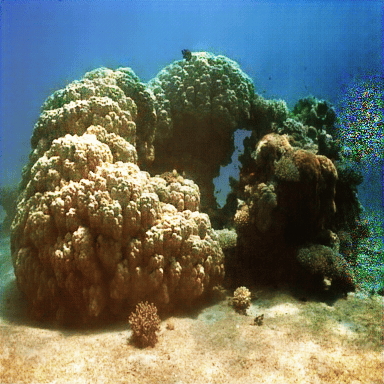}
& \img{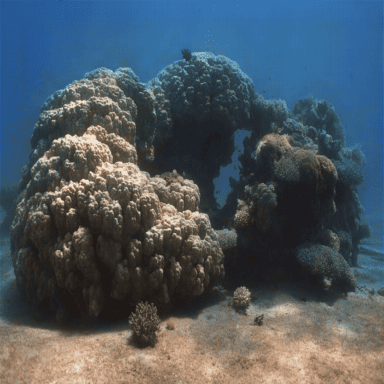}
& \img{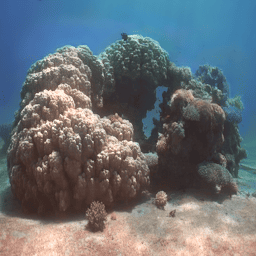}
\\

\lbl{SUIEB}
& \img{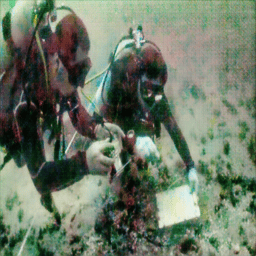}
& \img{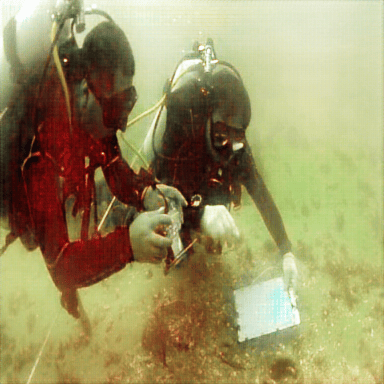}
& \img{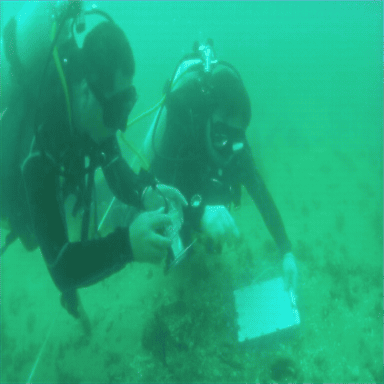}
& \img{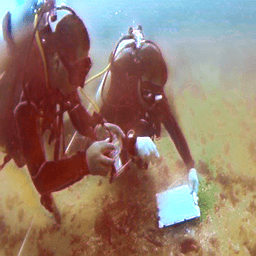}

& \img{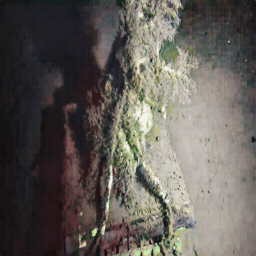}
& \img{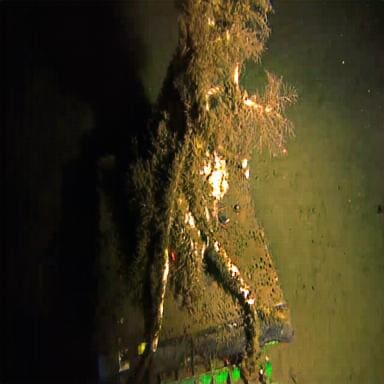}
& \img{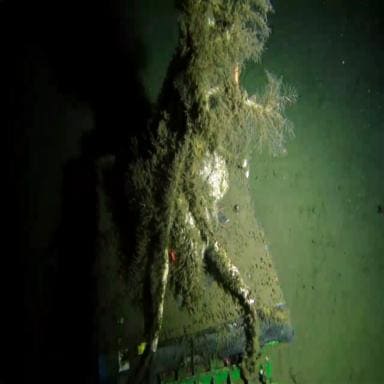}
& \img{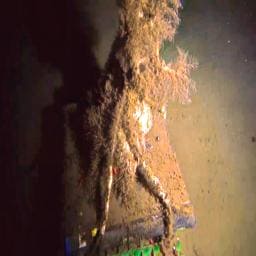}

& \img{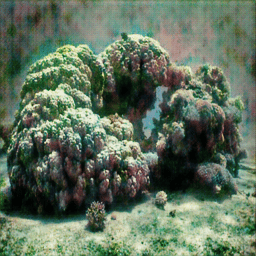}
& \img{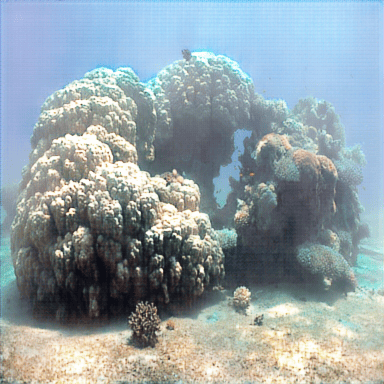}
& \img{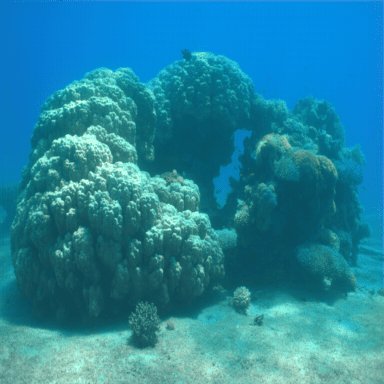}
& \img{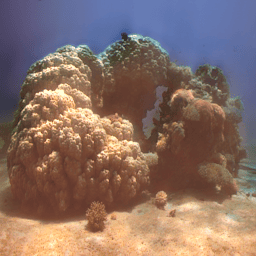}
\\

\lbl{PHISWID}
& \img{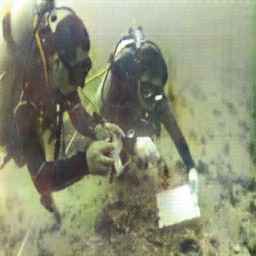}
& \img{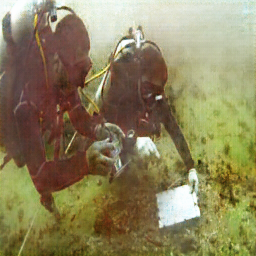}
& \img{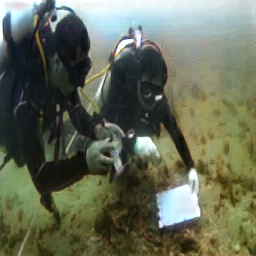}
& \img{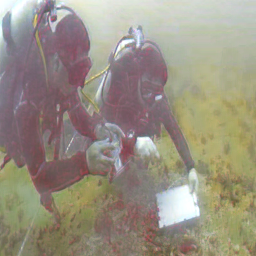}

& \img{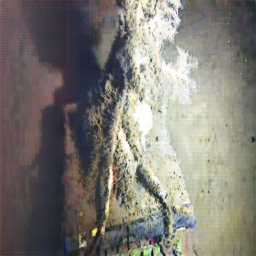}
& \img{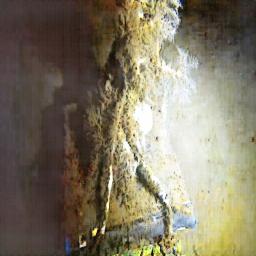}
& \img{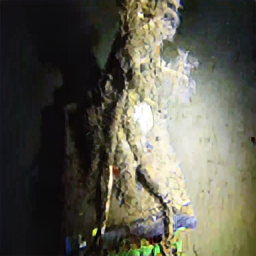}
& \img{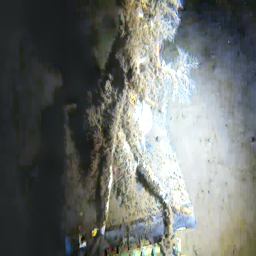}

& \img{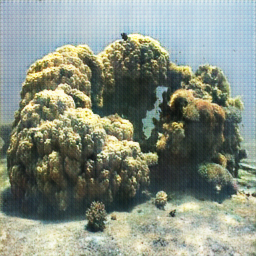}
& \img{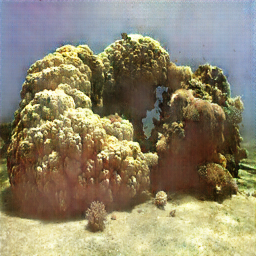}
& \img{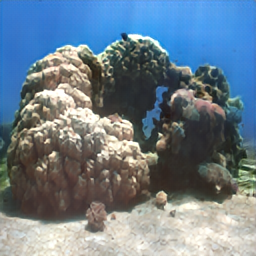}
& \img{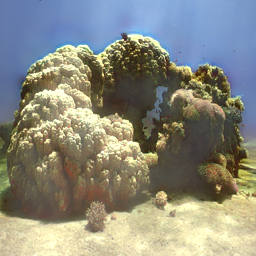}
\\

\lbl{Syrea}
& \img{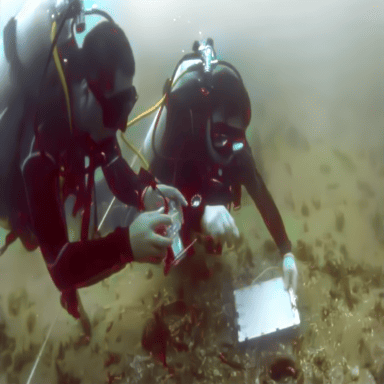}
& \img{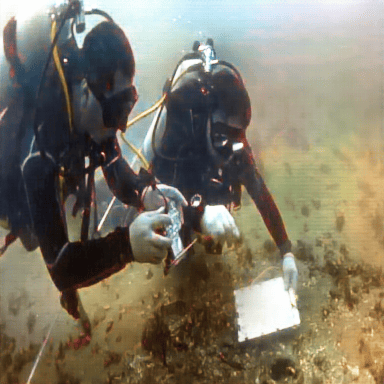}
& \img{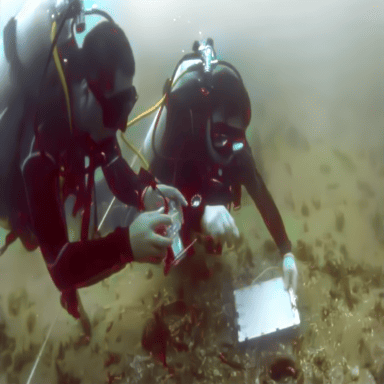}
& \img{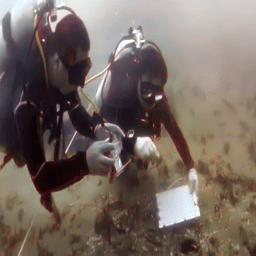}

& \img{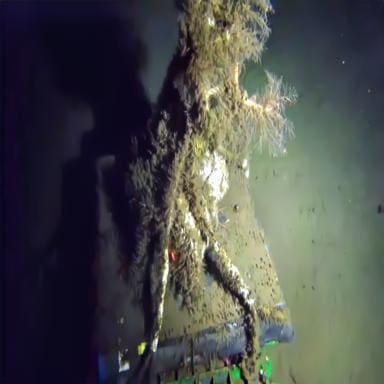}
& \img{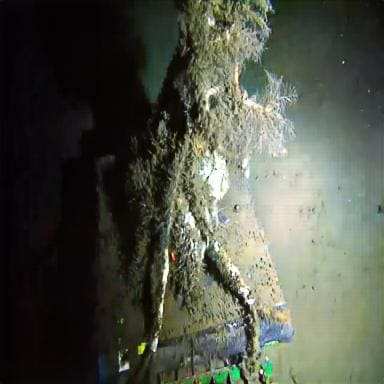}
& \img{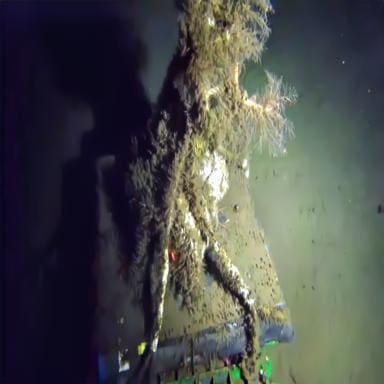}
& \img{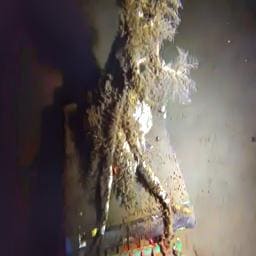}

& \img{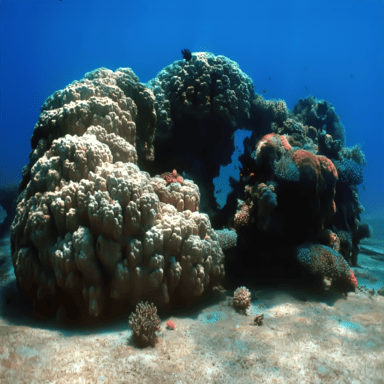}
& \img{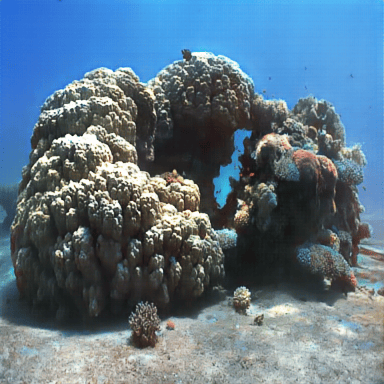}
& \img{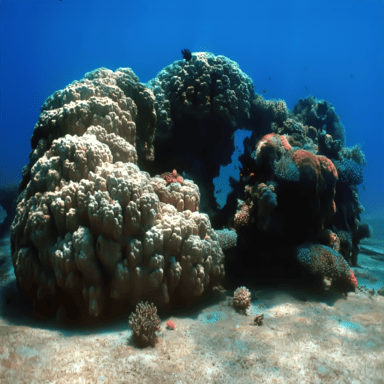}
& \img{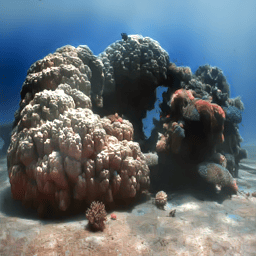}
\\

\lbl{\textbf{$\pi$-SUB}}

& \img{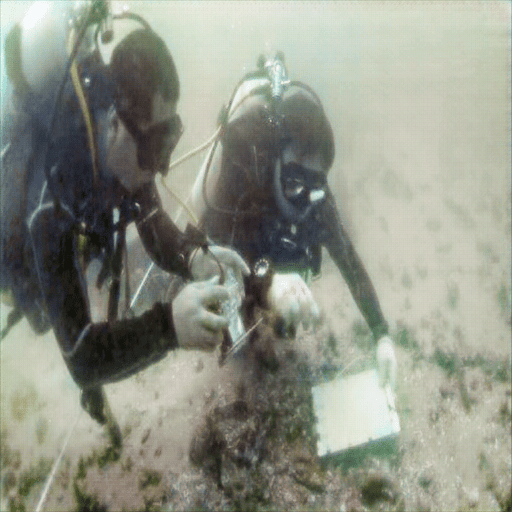}
& \img{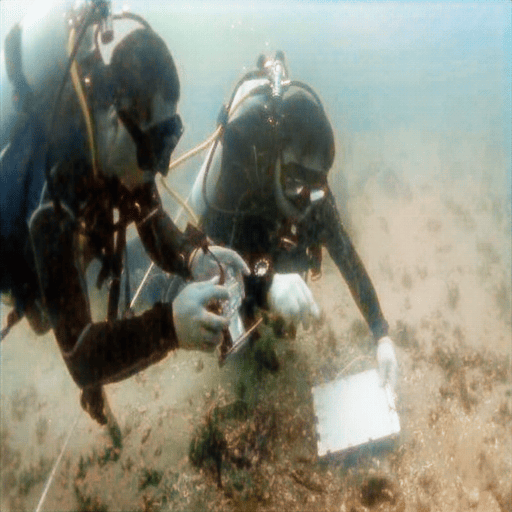}
& \img{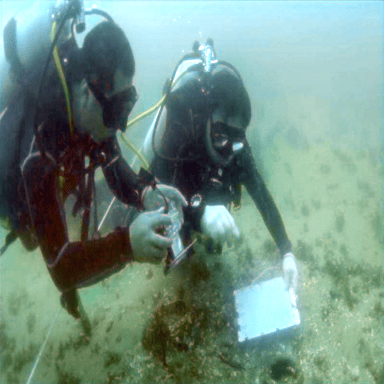}
& \img{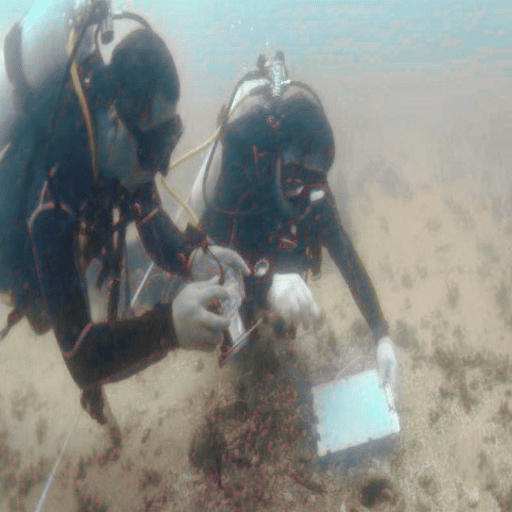}

& \img{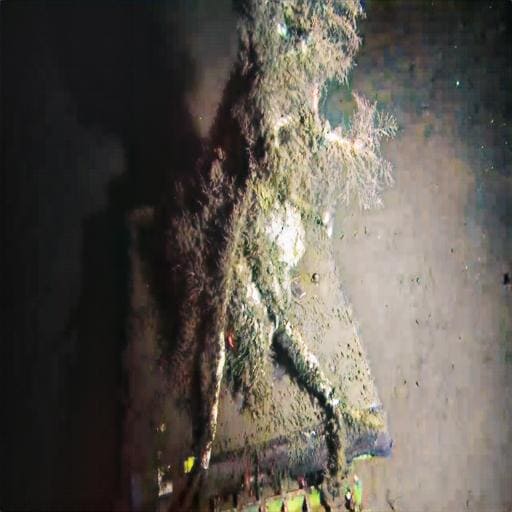}
& \img{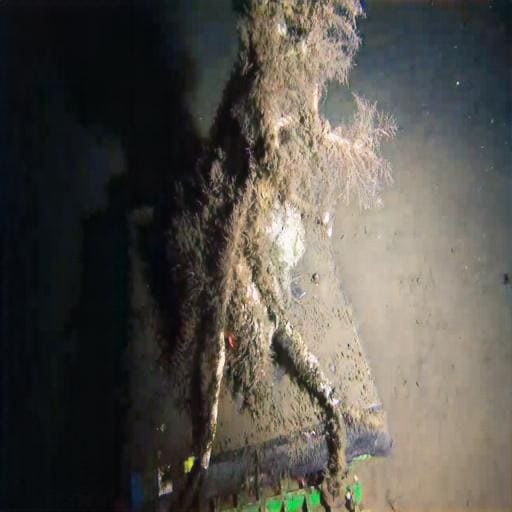}
& \img{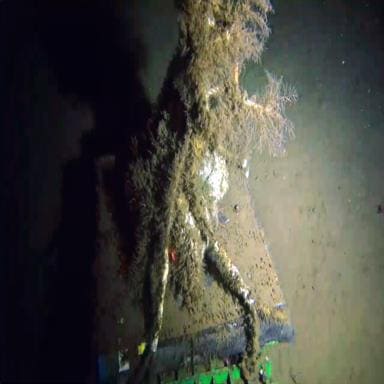}
& \img{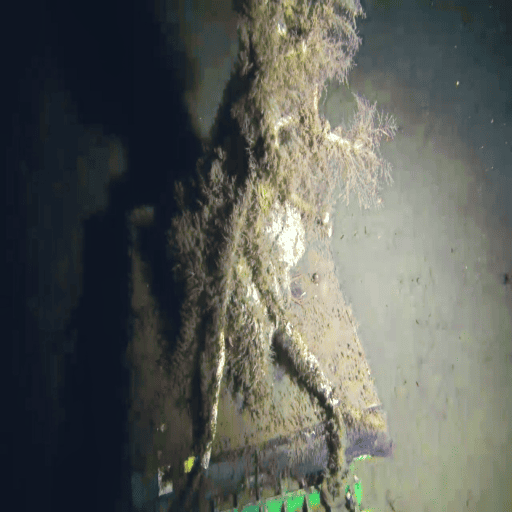}

& \img{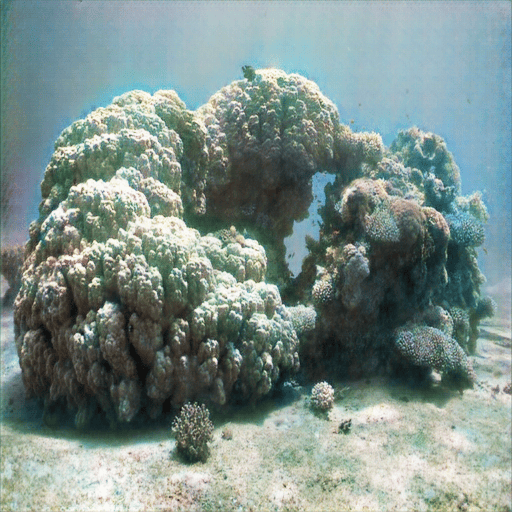}
& \img{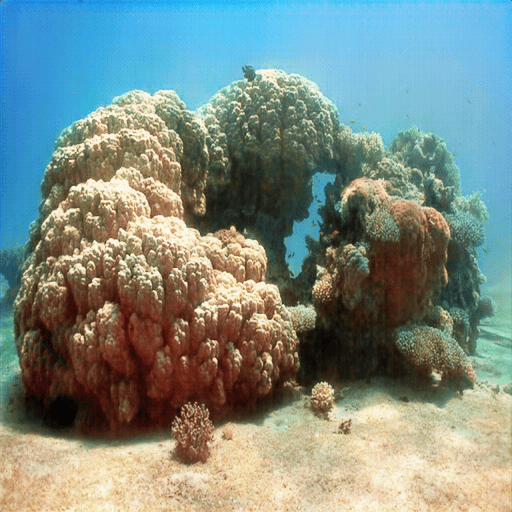}
& \img{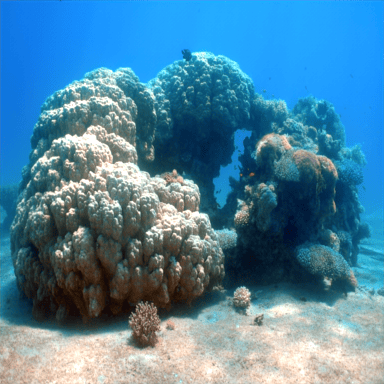}
& \img{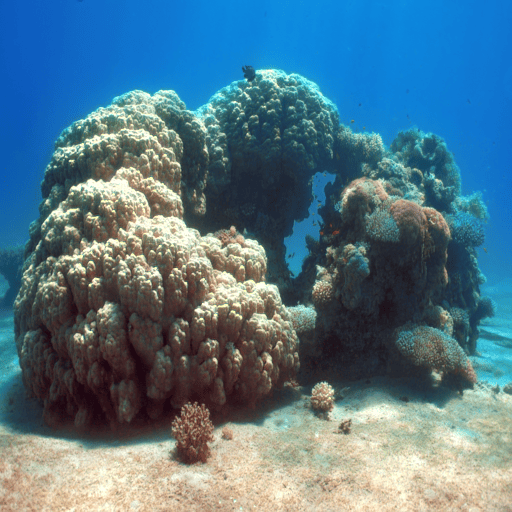}
\\

\end{tabular}

\caption{Representative qualitative enhancement results on three real benchmarks: UIEB (mixed shallow-water scenes), OceanDark (low-light deep-sea scenes), and SQUID (clear oceanic scenes). Columns (a)–(d) correspond to Pix2Pix, FUnIE-GAN, Phaseformer, and PUIE-Net, respectively.}

\label{fig:qualitative_results}
\end{figure*}

Figure~\ref{fig:qualitative_results} compares all four architectures on UIEB, OceanDark, and SQUID when trained on the real paired dataset UIEB, on the four competing synthetic datasets, and on $\pi$-SUB. Complete qualitative comparisons on all six benchmarks are provided in the supplementary material, and the observations below hold across them.

Each training source imparts a recognizable character to the enhanced outputs. Models trained on UIEB perform strongly on UIEB and U45, benefiting from the overlap between training and evaluation distributions, but transfer poorly to RUIE, OceanDark, SQUID, and FishTrac, consistent with close adaptation to the colour and haze character of UIEB's human-selected references. SUID-trained models show the strongest deviations, pushing bright regions toward saturation and mid-tones toward reddish-orange. SUIEB-trained models are inconsistent, sometimes leaving the raw cast intact and sometimes introducing warm casts of their own. PHISWID-trained models systematically under-correct, remaining washed-out and low in contrast, and on the yellow-dominant U45 scene PUIE-Net trained on PHISWID produces severe structural artifacts, indicating that its degradation model does not span this portion of the real colour-cast distribution. Syrea-trained models achieve balanced global correction but flatten chromatic content toward grey and appear over-smoothed. Models trained on $\pi$-SUB restore natural, saturated colour across all six benchmarks while preserving fine structure and object boundaries; the one caveat is a mildly warm tint on the blue-dominant SQUID and FishTrac scenes when paired with FUnIE-GAN, which the other three architectures do not exhibit.

\paragraph{\textbf{Quantitative Analysis}}
Table~\ref{tab:uiqm_niqe} reports UIQM and NIQE for all twenty-four trained models on UIEB, OceanDark, and SQUID. The supplementary material has the full table across all six benchmarks.

\begin{table}[h]
\renewcommand{\arraystretch}{1.2}
\caption{UIQM ($\uparrow$) and NIQE ($\downarrow$) of enhancement models trained on the six candidate training datasets and evaluated on UIEB, OceanDark (OD), and SQUID. Within each architecture group, the best and second-best score per benchmark and metric are shaded \colorbox{best}{best} and \colorbox{second}{second}.}
\label{tab:uiqm_niqe}
\footnotesize
\setlength{\tabcolsep}{3pt}
\resizebox{\columnwidth}{!}{
\begin{tabular}{@{}l|cc|cc|cc@{}}
\toprule
\multirow{2}{*}{Method} & \multicolumn{2}{c}{UIEB} & \multicolumn{2}{c}{OD} & \multicolumn{2}{c}{SQUID} \\
& UIQM$\uparrow$ & NIQE$\downarrow$ & UIQM$\uparrow$ & NIQE$\downarrow$ & UIQM$\uparrow$ & NIQE$\downarrow$ \\
\midrule
Pix2Pix@UIEB & 3.2702 & 7.4912 & 3.0191 & 8.5609 & 2.7705 & 8.9478 \\
Pix2Pix@SUID & 3.1284 & 5.3240 & 2.8849 & 7.9288 & 2.8128 & 8.9778 \\
Pix2Pix@SUIEB & 3.2587 & 5.7889 & \cellcolor{second}3.3383 & 6.2387 & 3.1036 & \cellcolor{second}6.1689 \\
Pix2Pix@PHISWID & \cellcolor{second}3.3577 & 6.6060 & 3.2570 & 6.6190 & 3.0979 & 8.5969 \\
Pix2Pix@Syrea & 3.2587 & \cellcolor{second}4.2789 & 3.2441 & \cellcolor{second}5.9893 & \cellcolor{second}3.2033 & 7.4289 \\
\textbf{Pix2Pix@$\pi$-SUB} & \cellcolor{best}\textbf{3.4129} & \cellcolor{best}\textbf{2.6666} & \cellcolor{best}\textbf{3.3495} & \cellcolor{best}\textbf{3.4896} & \cellcolor{best}\textbf{3.2777} & \cellcolor{best}\textbf{3.1016} \\
\midrule
FUnIE-GAN@UIEB & \cellcolor{best}3.2630 & 6.8676 & 2.7247 & 7.8561 & 2.4233 & 7.8850 \\
FUnIE-GAN@SUID & 3.2028 & 4.5057 & 2.6783 & \cellcolor{second}4.8080 & \cellcolor{second}3.0792 & 5.4466 \\
FUnIE-GAN@SUIEB & 3.2075 & 5.9874 & 2.7160 & 5.0544 & 2.7825 & 5.0768 \\
FUnIE-GAN@PHISWID & 3.2169 & 6.2559 & \cellcolor{best}2.9518 & 6.5030 & 3.0582 & 7.6149 \\
FUnIE-GAN@Syrea & 3.2345 & \cellcolor{second}3.8422 & 2.5171 & 5.6667 & 3.0516 & \cellcolor{second}4.7502 \\
\textbf{FUnIE-GAN@$\pi$-SUB} & \cellcolor{second}\textbf{3.2393} & \cellcolor{best}\textbf{2.4807} & \cellcolor{second}\textbf{2.8874} & \cellcolor{best}\textbf{3.9624} & \cellcolor{best}\textbf{3.1080} & \cellcolor{best}\textbf{3.4308} \\
\midrule
Phaseformer@UIEB & 3.0298 & 6.4296 & 2.4974 & 7.8588 & 1.9655 & 7.3066 \\
Phaseformer@SUID & 2.9828 & 6.9559 & 2.6325 & 5.3154 & 2.3539 & 7.1548 \\
Phaseformer@SUIEB & 2.6954 & 5.4515 & 2.3843 & 4.4672 & 1.7822 & 5.0577 \\
Phaseformer@PHISWID & \cellcolor{second}3.1306 & 8.4086 & \cellcolor{second}2.6996 & 6.1256 & \cellcolor{best}2.9865 & 7.1385 \\
Phaseformer@Syrea & 2.8950 & \cellcolor{second}5.2779 & 2.2358 & \cellcolor{second}4.4180 & 2.3305 & \cellcolor{best}3.0397 \\
\textbf{Phaseformer@$\pi$-SUB} & \cellcolor{best}\textbf{3.2329} & \cellcolor{best}\textbf{2.8519} & \cellcolor{best}\textbf{2.7367} & \cellcolor{best}\textbf{3.9225} & \cellcolor{second}\textbf{2.7771} & \cellcolor{second}\textbf{3.0414} \\
\midrule
PUIE-Net@UIEB & 3.0821 & 6.3810 & 2.6740 & 6.9020 & 2.5961 & 6.8937 \\
PUIE-Net@SUID & 3.1358 & 5.1542 & 2.6414 & 7.4799 & 2.7530 & 7.8411 \\
PUIE-Net@SUIEB & \cellcolor{best}3.2987 & 5.7597 & 2.8244 & \cellcolor{second}5.1229 & 2.7265 & 7.8757 \\
PUIE-Net@PHISWID & 3.0735 & 6.0496 & \cellcolor{second}2.8551 & 6.8193 & \cellcolor{second}2.7862 & 6.5016 \\
PUIE-Net@Syrea & 3.0412 & \cellcolor{second}3.2931 & 2.5540 & 5.1417 & 2.4545 & \cellcolor{second}3.4849 \\
\textbf{PUIE-Net@$\pi$-SUB} & \cellcolor{second}\textbf{3.2269} & \cellcolor{best}\textbf{2.6860} & \cellcolor{best}\textbf{2.8579} & \cellcolor{best}\textbf{2.9919} & \cellcolor{best}\textbf{2.8298} & \cellcolor{best}\textbf{3.1233} \\
\bottomrule
\end{tabular}}
\end{table}

Models trained on $\pi$-SUB rank first or second in UIQM on UIEB, OceanDark, and SQUID in all twentyfour architecture--benchmark combinations shown, the best win rate of any candidate training set on these three benchmarks. For Pix2Pix, $\pi$-SUB is best on all three, with the largest margins on OceanDark ($16.1\%$ over SUID) and SQUID ($16.5\%$ over SUID). FUnIE-GAN is the setting in which baselines remain most competitive, with $\pi$-SUB best on SQUID and second on UIEB and OceanDark. The corresponding margins on U45, RUIE, and FishTrac are reported in the supplementary material and follow the same pattern. UIEB-trained models perform competitively with $\pi$-SUB on the in-distribution UIEB benchmark, reflecting their ability to reproduce the appearance characteristics of the training data. However, their performance degrades substantially on out-of-distribution datasets such as SQUID, which exhibits different water types, illumination conditions, and optical characteristics. For example, the $\pi$-SUB-trained FUnIE-GAN achieves a 28.3\% higher UIQM on SQUID, indicating that the enhancement learned from UIEB is largely dataset-specific rather than physically generalizable. In contrast, the greater optical diversity and physically grounded degradation modeled by $\pi$-SUB enable the learned enhancement to transfer more effectively across unseen underwater environments. Finally, averaged across four UIE architectures and six real benchmark datasets, models trained on $\pi$-SUB improve UIQM by 9.46\% over two strong competing synthetic datasets Syrea and 4.18\% over PHISWID, while reducing NIQE by 23.98\% and 48.78\%, respectively.

\begin{table*}[t]
\centering

    \renewcommand{\arraystretch}{1.2}
\caption{Underwater image enhancement performance of state-of-the-art methods and models trained on $\pi$-SUB, in terms of UIQM ($\uparrow$) and NIQE ($\downarrow$). The best and second-best score per benchmark and metric are shaded \colorbox{best}{best} and \colorbox{second}{second}.}
\label{tab:uie_comparison}
\resizebox{\textwidth}{!}{
\begin{tabular}{@{}lcccccccccccc@{}}

\hline
\textbf{Method} &
\multicolumn{2}{c}{\textbf{UIEB}} &
\multicolumn{2}{c}{\textbf{U45}} &
\multicolumn{2}{c}{\textbf{RUIE}} &
\multicolumn{2}{c}{\textbf{FishTrac}} &
\multicolumn{2}{c}{\textbf{OceanDark}} &
\multicolumn{2}{c}{\textbf{SQUID}} \\

& UIQM$\uparrow$ & NIQE$\downarrow$
& UIQM$\uparrow$ & NIQE$\downarrow$
& UIQM$\uparrow$ & NIQE$\downarrow$
& UIQM$\uparrow$ & NIQE$\downarrow$
& UIQM$\uparrow$ & NIQE$\downarrow$
& UIQM$\uparrow$ & NIQE$\downarrow$ \\
\hline

ULAP~\cite{song2018}
& 1.3230 & 7.3794
& 2.2044 & 6.3523
& 2.2723 & 5.5231
& 1.3429 & 6.5523
& 1.8888 & 4.8517
& 0.8095 & 5.2161 \\

P2CNet~\cite{rao2023}
& 2.5195 & 7.8159
& 3.1237 & 6.2819
& 2.9362 & 7.4776
& 2.3442 & 11.3184
& 2.7917 & 8.6685
& 2.4927 & 8.0566 \\

Phaseformer~\cite{khan2025phase}
& 2.5371 & 9.0984
& 3.0136 & 10.0297
& 2.1309 & 7.9462
& 1.3561 & 6.9397
& 2.1360 & 6.5882
& 2.2675 & 9.5063 \\

FUnIE-GAN~\cite{islam2020fast}
& 3.0493 & 8.1176
& 2.7083 & 9.0159
& 3.0465 & 9.3881
& 2.8346 & 8.1613
& 2.9723 & 7.6801
& 2.1556 & 6.8013 \\

PUIE-Net~\cite{Fu_2022}
& 3.1128 & 6.9152
& 2.6778 & 7.9918
& 3.0292 & 7.7949
& 2.4554 & 6.4421
& 2.9079 & 8.1561
& 2.1935 & 7.7716 \\

Pix2Pix~\cite{Isola2017pix2pix}
&\cellcolor{second}{\textbf{3.2702}} & 7.4912
& 3.2145 & 8.9644
& 2.9561 & 7.7668
& 3.0552 & 8.8543
&\cellcolor{second}{\textbf{3.0191}} & 8.5609
& 2.7705 & 8.9478 \\

UShape~\cite{peng2023lsui}
& 2.7857 & 7.4041
& 3.1912 & 7.2647
& 2.9351 & 7.4888
& 1.8180 & 10.7056
& 2.9122 & 7.8756
& 2.3135 & 8.9676 \\

Spectroformer~\cite{khan2024spectroformer}
& 2.4380 & 7.0620
& 2.9508 & 7.4322
& 2.8367 & 6.2236
& 2.9669 & 4.9601
& 2.5432 & 5.1699
& 2.3157 & 4.4353 \\

UDNet~\cite{saleh2025adaptive}
& 2.3913 & 7.5062
& 3.1420 & 7.1917
& 2.9282 & 6.6007
& 2.7387 & 6.6421
& 2.0922 & 4.9538
& 1.9418 & 7.3046 \\

WaterNet~\cite{li2019underwater}
& 2.5067 & 7.4418
& 3.2734 & 8.7466
& 3.0550 & 6.2991
& 2.1123 & 5.2207
& 2.3290 & 4.8337
& 2.0060 & 6.4154 \\

Pix2Pix@MUSE~\cite{li2025realistic}
& - & -
& 3.0279 & -
& 3.0726 & -
& - & -
& 2.7836 & -
& - & - \\

Pix2Pix@$\pi$-SUB
&\cellcolor{best}{\textbf{3.4129}} &\cellcolor{second}{\textbf{2.6666}}
& 3.2898 &\cellcolor{best}{\textbf{4.6579}}
&\cellcolor{best}{\textbf{3.3081}} &\cellcolor{best}{\textbf{3.2704}}
&\cellcolor{best}{\textbf{3.4383}} &\cellcolor{best}{\textbf{3.3387}}
&\cellcolor{best}{\textbf{3.3495}} &\cellcolor{second}{\textbf{3.4896}}
&\cellcolor{best}{\textbf{3.2777}} &\cellcolor{second}{\textbf{3.1016}} \\

FUnIE-GAN@$\pi$-SUB
& 3.2393 &\cellcolor{best}{\textbf{2.4807}}
& 3.3986 & 5.3609
& 3.0059 & 3.3553
&\cellcolor{second}{\textbf{3.4147}} &\cellcolor{second}{\textbf{3.7558}}
& 2.8874 & 3.9624
&\cellcolor{second}{\textbf{3.1080}} & \textbf{3.4308} \\

Phaseformer@$\pi$-SUB
& 3.2329 & 2.8519
&\cellcolor{best}{\textbf{3.4655}} &\cellcolor{second}{\textbf{5.2308}}
& 2.9786 &\cellcolor{second}{\textbf{3.3232}}
& 3.2278 & 4.0125
& 2.7367 & 3.9225
& 2.7771 &\cellcolor{best}{\textbf{3.0414}} \\

PUIE-Net@$\pi$-SUB
& 3.2269 & 2.6860
&\cellcolor{second}{\textbf{3.4183}} & 6.0644
&\cellcolor{second}{\textbf{3.1082}} & 4.1418
& 3.3990 & 3.8792
& 2.8579 &\cellcolor{best}{\textbf{2.9919}}
& 2.8298 & 3.1233 \\
\hline
\end{tabular}}
\end{table*}

Table~\ref{tab:uie_comparison} compares the $\pi$-SUB-trained models against a classical prior-based method (ULAP) and state-of-the-art learned enhancement models. A $\pi$-SUB-trained model achieves the highest UIQM on every benchmark, with margins of $4.4\%$ on UIEB, $5.9\%$ on U45, $7.7\%$ on RUIE, $10.9\%$ on OceanDark, $12.5\%$ on FishTrac, and $18.3\%$ on SQUID over the strongest competing method in each case. The NIQE margins are larger still, from $21.8\%$ on U45 to $64.1\%$ on UIEB, where FUnIE-GAN@$\pi$-SUB reduces the best baseline of 6.9152 to 2.4807. ULAP and the transformer-based Spectroformer remain competitive on individual benchmarks, but the $\pi$-SUB-trained models dominate both metrics overall. Because these gains persist across four architectures with distinct inductive biases, they are attributable to the training distribution rather than to any single model design. 

Further, statistical analysis using the Friedman rank test with Nemenyi post-hoc comparison across all four architectures and six benchmarks confirms that $\pi$-SUB yields significantly better mean ranks on both UIQM and NIQE. On the challenging U45 subset, Phaseformer trained on $\pi$-SUB further exhibits the lowest variance (UIQM 10.92 \%, NIQE 18.89 \%) compared with Syrea (14.60\% and 26.44\%) and PHISWID (12.31\% and 35.34\%), indicating greater robustness. Full details are given in the supplementary material.

\subsubsection{\textbf{Ablation Study}}
\begin{figure}[!t]
\centering
\setlength{\tabcolsep}{0.5pt}
\renewcommand{\arraystretch}{0.9}

\begin{tabular}{ccccc}
\scriptsize Input &
\scriptsize Base &
\scriptsize +Haze &
\scriptsize +Bio &
\scriptsize $\pi$-SUB \\

\includegraphics[width=0.194\linewidth]{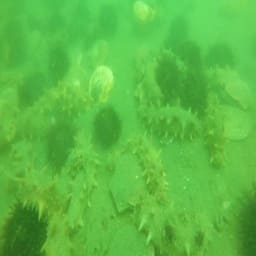} &
\includegraphics[width=0.194\linewidth]{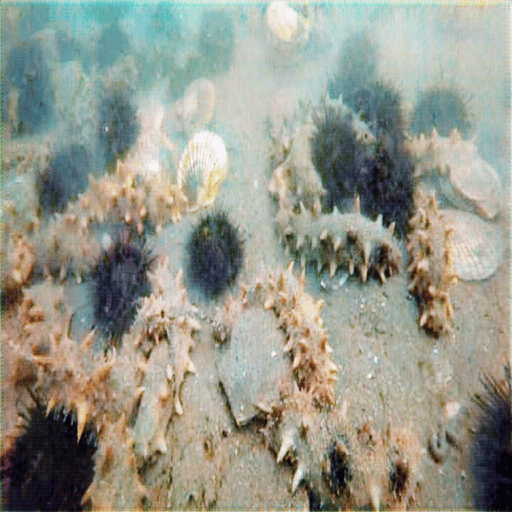} &
\includegraphics[width=0.194\linewidth]{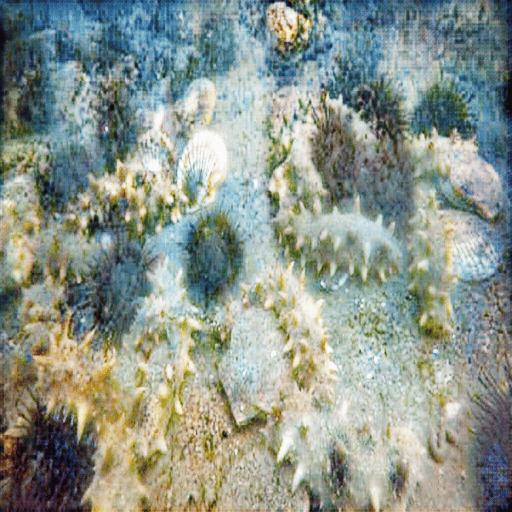} &
\includegraphics[width=0.194\linewidth]{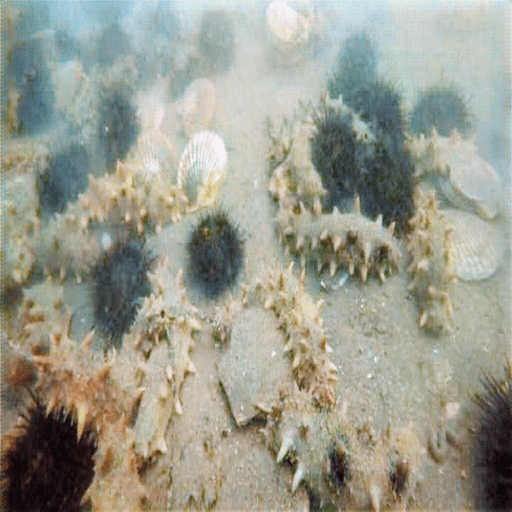} &
\includegraphics[width=0.194\linewidth]{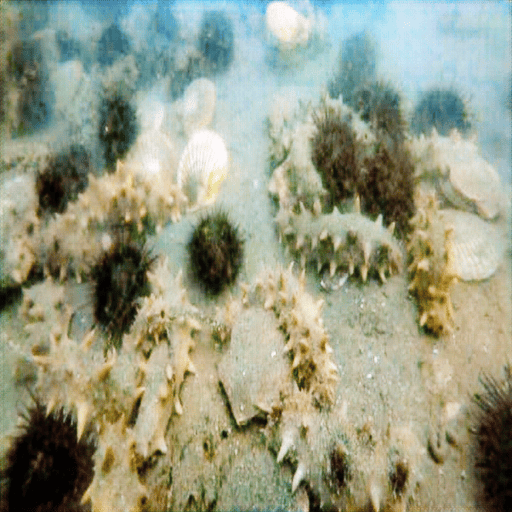} \\[-1mm]

\includegraphics[width=0.194\linewidth]{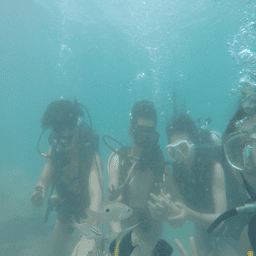} &
\includegraphics[width=0.194\linewidth]{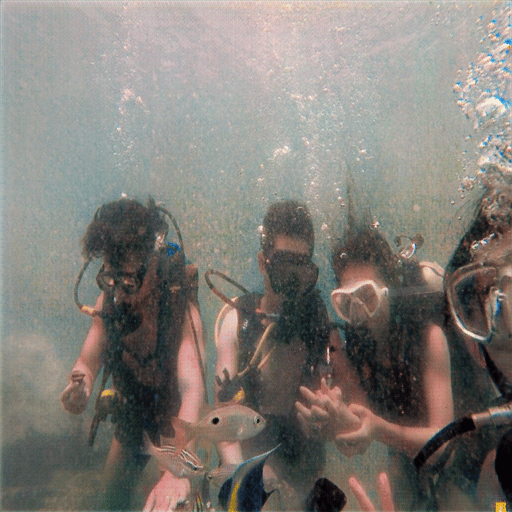} &
\includegraphics[width=0.194\linewidth]{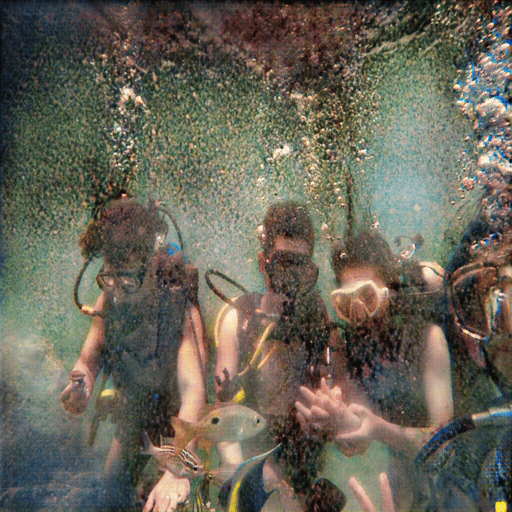} &
\includegraphics[width=0.194\linewidth]{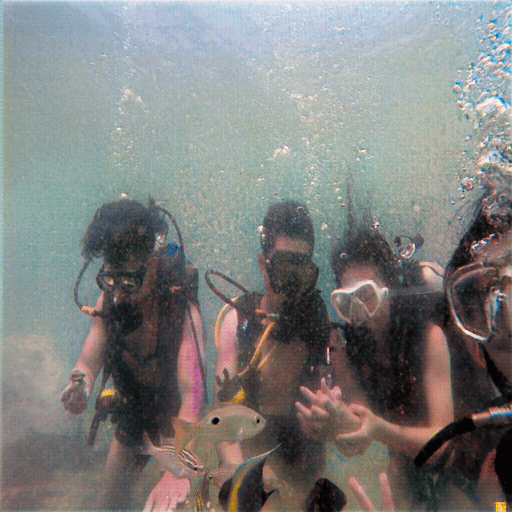} &
\includegraphics[width=0.194\linewidth]{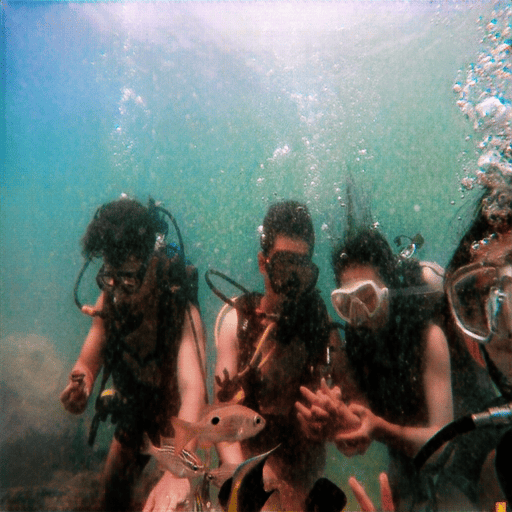} \\[-1mm]

\includegraphics[width=0.194\linewidth]{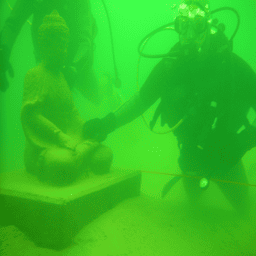} &
\includegraphics[width=0.194\linewidth]{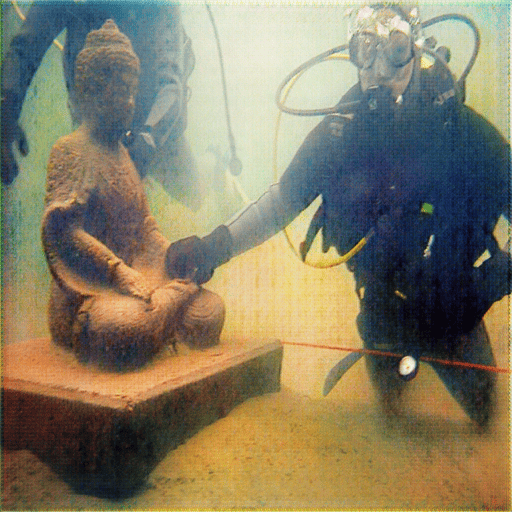} &
\includegraphics[width=0.194\linewidth]{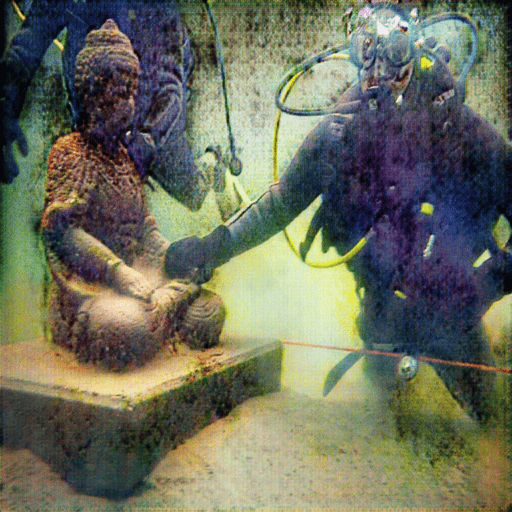} &
\includegraphics[width=0.194\linewidth]{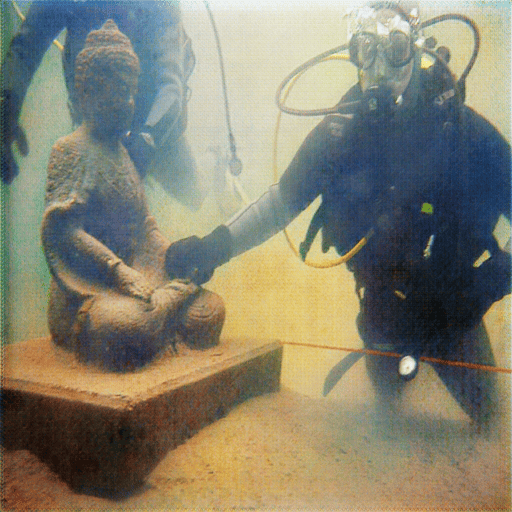} &
\includegraphics[width=0.194\linewidth]{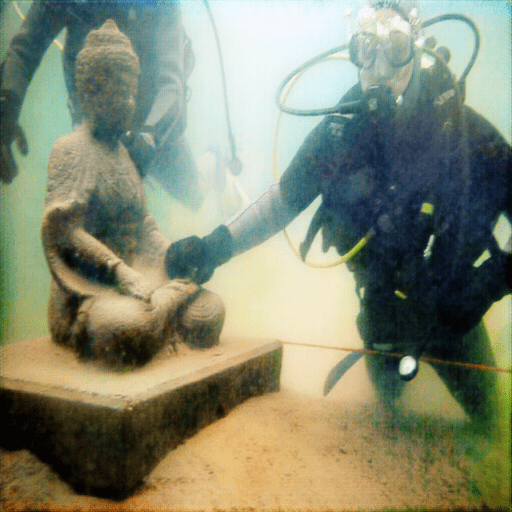}
\end{tabular}

\caption{Qualitative ablation of FUnIE-GAN under $\pi$-SUB training datasets.}
\label{fig:ablation}
\end{figure}

To verify that each residual phenomenon in the Augmented Realism stage contributes to generalization, five training variants of $\pi$-SUB are constructed: \textit{Base} (the deterministic MUIF subset $I_b$ only), \textit{Base+Bio} ($I_b$ with biological effects), \textit{Base+Haze} ($I_b$ with volumetric haze), \textit{Base+SP} ($I_b$ with suspended particulate matter), and \textit{Full} (all three residual phenomena combined). FUnIE-GAN is retrained from scratch on each variant under identical hyperparameters and evaluated on the six real benchmarks. Figure~\ref{fig:ablation} illustrates the contribution of each augmentation component to underwater image enhancement. The complete $\pi$-SUB configuration, combining all augmentation categories, consistently produces the best enhancement performance. This trend is confirmed quantitatively in Table~\ref{tab:ablation_metrics}, where the UIQM and NIQE scores, averaged across four UIE architectures and six real-world underwater benchmark datasets, consistently favor the complete $\pi$-SUB configuration. 

\begin{table}[htbp]
    \centering
    \caption{Ablation study of $\pi$-SUB across six real-world benchmarks.}
    \label{tab:ablation_metrics}
    \small
    \setlength{\tabcolsep}{12pt}
    \begin{tabular}{lcc}
        \toprule
        Variant & UIQM$\uparrow$ & NIQE$\downarrow$ \\
        \midrule
        Base           & 2.8713 & 4.7283 \\
        Base+Bio       &3.1138 & 3.7961 \\
        Base+Haze      & 3.0941 & 3.9254  \\
        Base+SP        & 2.9851 & 4.3382  \\
        \textbf{Full}  &\cellcolor{best}{\textbf{3.1829}} & \cellcolor{best}{\textbf{3.4129}} \\
    \bottomrule
    \end{tabular}
\end{table}

Removing any single residual phenomenon from the training distribution degrades performance relative to the full configuration. Biological effects alone improve warm-wavelength recovery in green-dominant scenes such as those in OceanDark, since biological absorption primarily attenuates the red and yellow bands, but they do not reconstruct the low-contrast haze artifacts of turbid open-water scenes in FishTrac and SQUID. Haze alone improves contrast but over-compensates in detail reconstruction. Only the \textit{Full} configuration achieves top performance consistently across benchmarks, confirming that the residual phenomena are complementary rather than redundant components of a physically complete training distribution.

\subsubsection{\textbf{Keypoint Feature Matching}}

The practical value of enhancement is ultimately reflected in downstream tasks such as visual odometry, SLAM, and 3D reconstruction, which depend on reliable feature correspondences across frames. Matching is evaluated with SIFT~\cite{lowe2004distinctive}, whose scale-space detector suits the low contrast, blur, and residual haze of underwater imagery; initial correspondences from Lowe's ratio test are refined by RANSAC-based fundamental matrix estimation~\cite{fischler1981random} to retain only geometrically consistent matches.

\begin{figure}[h]
    \setlength{\tabcolsep}{2pt}
    \renewcommand{\arraystretch}{0.95}
    \begin{tabular}{ccc}
        \includegraphics[width=0.32\linewidth, height=0.23\linewidth]{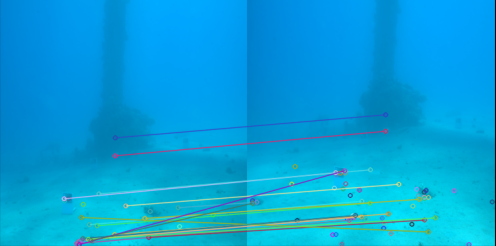} &
        \includegraphics[width=0.32\linewidth, height=0.23\linewidth]{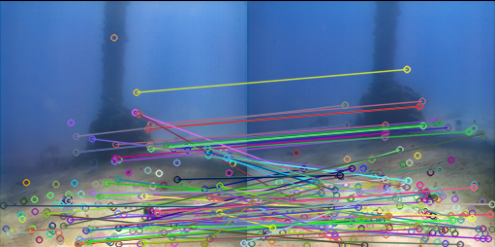} &
        \includegraphics[width=0.32\linewidth, height=0.23\linewidth]{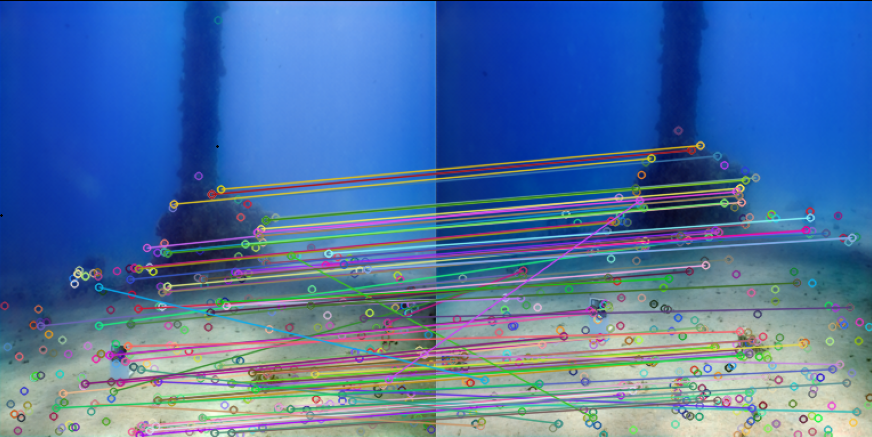} \\
        \small{Raw (23)} & \small{PHISWID (85)} & \small{$\pi$-SUB (157)} \\
    \end{tabular}
    \caption{SIFT keypoint matching between consecutive SQUID Katzaa dataset}
    \label{fig:keypoint_matching}
\end{figure}

On consecutive SQUID Katzaa frames as shown in Figure~\ref{fig:keypoint_matching}, the raw pair yields only 23 verified matches. PhaseFormer trained on PHISWID~\cite{kaneko2026phiswid} raises this to 85, while training on $\pi$-SUB yields 157: a $6.8\times$ improvement over the raw frames and $1.8\times$ over PHISWID. The substantially higher count of verified correspondences shows that $\pi$-SUB better preserves the local image structure required for feature-dependent downstream applications.

In summary, the results validate the $\pi$-SUB framework from two complementary perspectives. Distributionally, the $\pi$-SUB dataset aligns more closely with real underwater imagery than existing synthetic datasets, achieving the lowest global and cluster-wise FID, an OOD rate of 2.00\%, and complete coverage of the seven perceptual regimes observed in real benchmarks. Functionally, models trained on $\pi$-SUB consistently generalize better across six real benchmark datasets, achieving superior UIQM and NIQE while preserving natural color and structural detail. In the downstream feature-matching task, $\pi$-SUB nearly doubles the number of geometrically consistent keypoint matches compared with PHISWID. These results demonstrate that the combination of measured Jerlov optical properties, depth-dependent irradiance, and biological spectral modulation produces a training distribution that improves both enhancement quality and downstream underwater perception.

\section{Conclusion}
\label{sec:con}

This paper presents a novel physics-informed synthetic underwater benchmark framework named $\pi$-SUB. Using the proposed framework, the $\pi$-SUB dataset comprising of paired synthetic underwater--reference images spanning shallow-to-deep and coastal-to-oceanic environments. The framework makes three principal contributions: (i) a physically consistent degradation pipeline that models depth-dependent downwelling irradiance, biological optical processes including chlorophyll-\emph{a} absorption, CDOM, and fluorescence, together with forward and backward environmental scattering across all ten Jerlov water types; (ii) a comprehensive paired benchmark accompanied by complete physical metadata; and (iii) a two-axis validation methodology based on \emph{hyper-realism} and \emph{generalizability}.

Hyper-realism was demonstrated through distributional analysis against the pooled distribution of five real-world benchmarks. $\pi$-SUB achieved a global FID of 95, a 46\% reduction relative to the best existing synthetic benchmark (176, Syrea; the remaining comparators being SUID, SUIEB, and PHISWID), an out-of-distribution rate of only 2.00\%, and was the only synthetic benchmark covering all seven perceptual regimes observed in real underwater imagery.

Generalizability was demonstrated using twenty-four enhancement models spanning generative, convolutional, and transformer-based architectures. Since Syrea and PHISWID are the strongest competing synthetic benchmarks in terms of realism and enhancement performance, they are used as the primary baselines. Averaged across four UIE architectures and six real benchmark datasets, models trained on $\pi$-SUB improve UIQM by 9.46\% over Syrea and 4.18\% over PHISWID, while reducing NIQE by 23.98\% and 48.78\%, respectively. In a downstream feature-matching task, $\pi$-SUB increases the number of geometrically consistent SIFT correspondences from 23 on the raw images to 157, representing a 1.8$\times$ improvement over PHISWID.

These results demonstrate that physically complete modeling of underwater optical variability as presented in $\pi$-SUB produces synthetic data that is both hyper-realistic and generalizable for underwater image enhancement. Future work will extend the framework toward artificially illuminated deep-sea environments, highly turbid coastal water, and  caustics that remain underrepresented in existing real-world benchmarks.

\balance

\end{document}